\documentclass{article}
\usepackage{iclr2027_conference,times}
\iclrfinalcopy
\usepackage{hyperref}
\hypersetup{hidelinks}
\usepackage{url}
\usepackage{graphicx}
\usepackage{wrapfig}
\usepackage{placeins}
\usepackage{algorithm}
\usepackage{algorithmic}
\usepackage{listings}
\usepackage{booktabs}
\usepackage{array}
\usepackage{amsmath}
\usepackage{amssymb}
\usepackage[table]{xcolor}
\usepackage{adjustbox}
\usepackage{multirow}
\usepackage{makecell}
\fancypagestyle{labpaperfirst}{
  \fancyhf{}
  \fancyhead[L]{\includegraphics[width=4.2in]{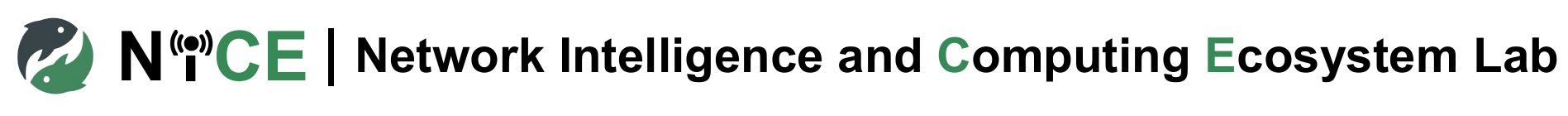}}
  \renewcommand{\headrulewidth}{0pt}
}
\newcolumntype{L}[1]{>{\raggedright\arraybackslash}p{#1}}
\definecolor{bestcell}{RGB}{226,239,218}
\definecolor{worstcell}{RGB}{244,204,204}
\definecolor{sourceconditioncolor}{RGB}{31,78,121}
\definecolor{acceptedconditioncolor}{RGB}{0,112,80}
\definecolor{rejectedconditioncolor}{RGB}{180,35,45}

\newcommand{\sourcecondition}[1]{%
  \colorbox{blue!12}{\textcolor{sourceconditioncolor}{\textbf{#1}}}}
\newcommand{\acceptedcondition}[1]{%
  \colorbox{green!15}{\textcolor{acceptedconditioncolor}{\textbf{#1}}}}
\newcommand{\rejectedcondition}[1]{%
  \colorbox{red!12}{\textcolor{rejectedconditioncolor}{\textbf{#1}}}}
\title{CROP: Task Relevance via Counterfactuals for Selective On-Policy Distillation}
\author{
\bfseries Enhan Li\textsuperscript{1},\enspace Junhao He\textsuperscript{1},\enspace Hongyang Du\textsuperscript{1},\thanks{Corresponding author.}\\[5pt]
{\normalfont\normalsize\textsuperscript{1}The University of Hong Kong}
}

\begin{document}

\maketitle
\thispagestyle{labpaperfirst}
\lhead{Preprint}
\renewcommand{\headrulewidth}{0.4pt}

\begin{abstract}

On-policy distillation (OPD) supervises a student language model on trajectories sampled from its current policy, but applies supervision uniformly across response positions.
Selective OPD addresses this limitation by allocating supervision non-uniformly across response tokens according to their estimated training value. Most existing criteria, however, focus primarily on \emph{optimization need}, such as uncertainty or teacher-student disagreement, while \emph{task relevance}, namely whether the supervision is tied to the semantic content of the current input, remains less directly characterized.
To address this gap, we introduce Counterfactual Relevance for On-Policy Distillation (CROP), which operationalizes task relevance through a paraphrase-calibrated counterfactual sensitivity margin.
For each source prompt, CROP constructs a validated original–paraphrase–counterfactual triplet, holds the student rollout fixed, and measures each response position by the teacher's sensitivity to a task-relevant condition change calibrated by its sensitivity to a meaning-preserving paraphrase.
At a 20\% per-response token budget, CROP attains the best aggregate score in two
teacher--student settings, exceeding the strongest non-CROP baseline by 1.43
and 0.96 points. Among matched controls, CROP gains 2.06 points over the
reversed-ranking CROP-Bottom, while replacing teacher with student rescoring
reduces the average score by 0.61 points. Across 5\%–20\% supervision budgets, CROP consistently outperforms the strongest matched hard selector, with its largest observed margin at the lowest 5\% budget.
\end{abstract}

\section{Introduction}

Post-training is a critical stage for adapting pretrained language models to
downstream objectives and improving their capabilities and controllability~\citep{ouyang2022training}.
Two dominant approaches are supervised fine-tuning (SFT) and reinforcement
learning (RL)~\citep{jiang2026sftvsrl}.
SFT learns from fixed offline demonstrations, whereas RL optimizes the policy
using rewards on sampled outputs~\citep{ouyang2022training,schulman2017ppo}.
A natural objective is therefore to retain dense supervision while adapting it to the states actually visited by the student. On-policy distillation (OPD) addresses this objective by querying a stronger teacher on
prefixes visited by the student's current policy, combining on-policy coverage
with dense token-level distributional supervision~\citep{yang2026learning,song2026survey}.
However, standard OPD treats every visited response token as an equally suitable
imitation target.
This uniform allocation can waste supervision on input-generic tokens while underemphasizing tokens that carry task-specific learning signals.

Selective OPD methods recognize that token value is heterogeneous, and recent
work has developed a broad range of token-level selection and weighting
strategies~\citep{song2026survey}.
Uncertainty-based selectors use student entropy or entropy-divergence
combinations to identify informative positions~\citep{jin2026eopd,xu2026tip}.
Other methods prioritize teacher-student disagreement that is locally
teachable or reliable~\citep{wang2026taopd,xing2026trust}.
Position- and training-dynamics-based methods use trajectory position,
persistent loss, or accumulated discrepancy as additional proxies~\citep{liu2026pwopsd,xie2026position,jiang2026rock}.
A further line constructs contrastive, advantage-based, or counterfactual
token-level credit using negative prompts, privileged-policy comparisons, or
sibling rollouts~\citep{shen2026credit,yu2026dual,meng2026craft}.
Most existing selectors nevertheless center on optimization need, asking whether a token is uncertain, differs from the teacher, is teachable or reliable, or is otherwise useful for training.

\begin{figure*}[t]
\centering
\includegraphics[width=0.90\textwidth]{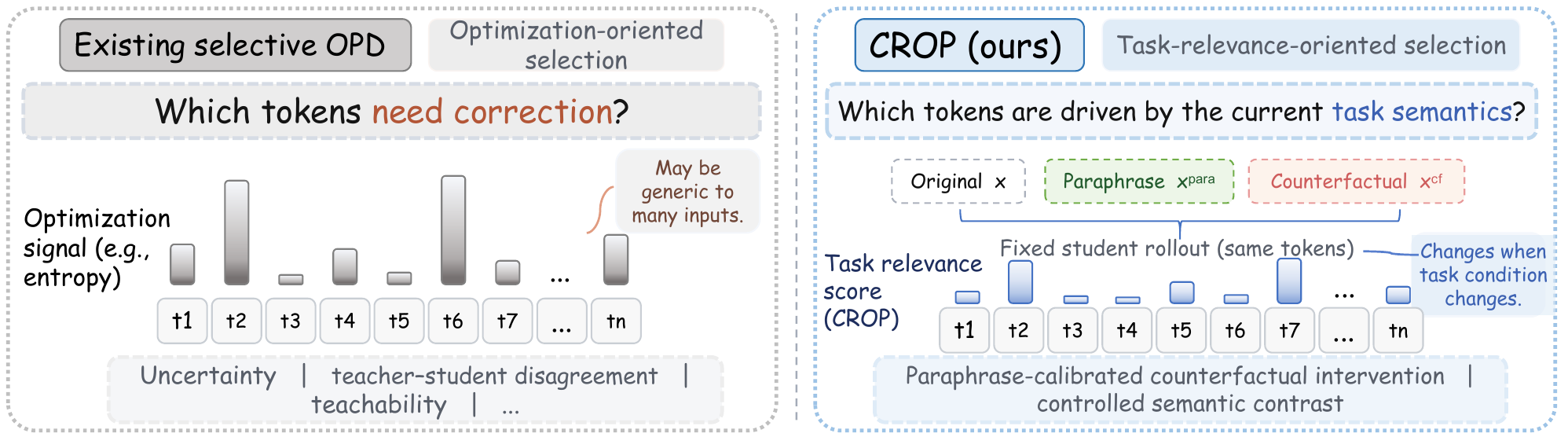}
\caption{Optimization-oriented versus task-relevance-oriented token selection in OPD. Existing selectors mainly identify tokens that may benefit from correction using signals such as uncertainty or teacher–student disagreement. CROP focuses on tokens whose supervision is sensitive to controlled changes in the task semantics.}
\label{fig:motivation}
\end{figure*}

However, optimization-oriented signals do not fully characterize a token's
supervision value.
For example, TIP shows that entropy alone can miss low-entropy but high-divergence positions, indicating that valuable supervision cannot be captured by a single optimization proxy~\citep{xu2026tip}. More fundamentally, even tokens that are uncertain, discrepant from the teacher, or readily correctable may still reflect input-generic response patterns, while low-entropy positions can remain task-relevant. This suggests a distinction between optimization need and task relevance that existing selective OPD methods do not directly resolve, as conceptually illustrated in Figure~\ref{fig:motivation}.

Optimization need asks whether a token may benefit from correction, whereas task relevance asks whether improving it teaches the student something specific to the semantic content of the current input. Along this direction,
CREDIT moves toward task-sensitive supervision by rescoring a fixed response and feedback under unrelated batch queries~\citep{shen2026credit}.
However, such unrelated prompts may differ in both task semantics and other input-level properties, making the resulting contrast difficult to attribute specifically to task-relevant variation. A natural solution is therefore to introduce a controlled counterfactual intervention. Specifically, if a token is genuinely tied to a task-relevant condition, changing that condition while holding the student rollout fixed should induce a corresponding change in the teacher’s supervision. We therefore introduce Counterfactual Relevance for On-Policy Distillation (CROP), which constructs a matched original-paraphrase-counterfactual triplet for each source prompt. The paraphrase preserves task-relevant meaning, while the counterfactual changes one material condition.
CROP holds the student's original rollout fixed, compares token distributions under the three prompts, and subtracts paraphrase sensitivity from counterfactual sensitivity.
The highest-scoring positions are then selected under a fixed per-response token budget
for the unchanged sampled-token OPD update, yielding a model-internal, contrast-specific measure of controlled semantic dependence.

Our contributions are:
\begin{itemize}
\item \textbf{Task relevance for selective OPD.}
We distinguish a token's task relevance from optimization need and motivate controlled semantic intervention as a direct way to characterize task-relevant supervision at the token level.
Building on this perspective, we use matched original-paraphrase-counterfactual prompts and a fixed realized student rollout to derive a model-internal, contrast-specific relevance signal.
\item \textbf{Paraphrase-calibrated counterfactual token selection.}
We propose CROP, which measures each response position through counterfactual sensitivity calibrated by a meaning-preserving paraphrase, and uses the resulting relevance scores to allocate a fixed per-response token budget while leaving the underlying OPD objective unchanged.
\item \textbf{Performance and token-selection evidence.}
At a 20\% per-response token budget, CROP attains the best aggregate score in both
teacher--student settings, leading the strongest non-CROP baseline by 1.43 and
0.96 points, respectively.
Across 5\%–20\% budgets, CROP consistently outperforms the strongest matched hard selector, with its largest margin at 5\%; notably, 5\% supervision reaches an average score of 26.47, only 0.51 points below the best 20\% result. Token-selection analysis further shows that CROP retains low-entropy positions largely missed by uncertainty-based selectors.
\end{itemize}

\section{Related Work}

\paragraph{On-policy distillation.}
OPD reduces the train-test state mismatch of offline distillation by querying a
teacher on prefixes visited by the current student~\citep{yang2026learning,song2026survey}.
Recent variants improve its objective,
stability, and use of teacher uncertainty~\citep{jin2026eopd,ko2026reopold,yang2026oglssd,li2026rethinking}.
A closely related line asks whether every teacher-supervised token should receive
equal weight. TIP combines entropy with teacher-student divergence~\citep{xu2026tip};
TA-OPD focuses on whether a correction is locally teachable~\citep{wang2026taopd};
TrOPD restricts supervision to teacher-reliable trust
regions~\citep{xing2026trust}; PW-OPSD and IW-OPD use trajectory position and
accumulated distributional discrepancy as reliability signals~\citep{liu2026pwopsd,xie2026position};
and Rock Tokens studies whether
persistently high-loss tokens are functionally necessary~\citep{jiang2026rock}.
These approaches expose
important dimensions of supervision quality. Their selectors rely mainly on
statistical, positional, or optimization proxies. 
Unlike these optimization-oriented selectors, CROP targets task relevance through controlled prompt interventions.

\paragraph{Credit assignment in reinforcement learning and distillation.}
Credit assignment determines how a coarse training signal is distributed across
the decisions that produced an outcome. In language-model post-training, token
selection and weighting have been used to refine sequence-level rewards,
preferences, and dense distillation targets. GRPO-style methods redistribute
outcome rewards across tokens. Selective distillation methods decide where
imitation should be applied.
CREDIT holds the response and feedback fixed and subtracts teacher-side scores under unrelated
inputs, yielding a prior-contrastive self-distillation reward~\citep{shen2026credit}.
DOPD instead routes supervision by privileged-policy advantage gaps~\citep{yu2026dual}.
Within this credit-assignment framing, CROP converts token-level relevance into a hard selection mask for external-teacher OPD.

\paragraph{Counterfactual reasoning in post-training.}
\citet{sharma2024sycophancy} show that preference judgments can favor
responses aligned with users' stated beliefs over truthful answers;
\citet{dubois2024length} estimate length-controlled preferences by conditioning
evaluator judgments on equal response lengths; and
\citet{chen2024odin} introduce ODIN, which separates length-correlated and
content-related reward components, using only the latter during RL.
Causal reward modeling further addresses spurious correlations:
\citet{wang2025beyond} enforce counterfactual invariance to irrelevant
attributes while preserving task-relevant content; and
\citet{kim2026mitigating} construct matched response pairs that hold content
or length fixed to separate content quality from verbosity.
Within this interventionist perspective, CROP calibrates a counterfactual prompt variation with a matched paraphrase to derive a token-level relevance signal for OPD.

\section{Method: Counterfactual Relevance for On-Policy Distillation}

\subsection{Overview and Problem Setup}

\begin{figure*}[t]
\centering
\includegraphics[width=0.95\textwidth]{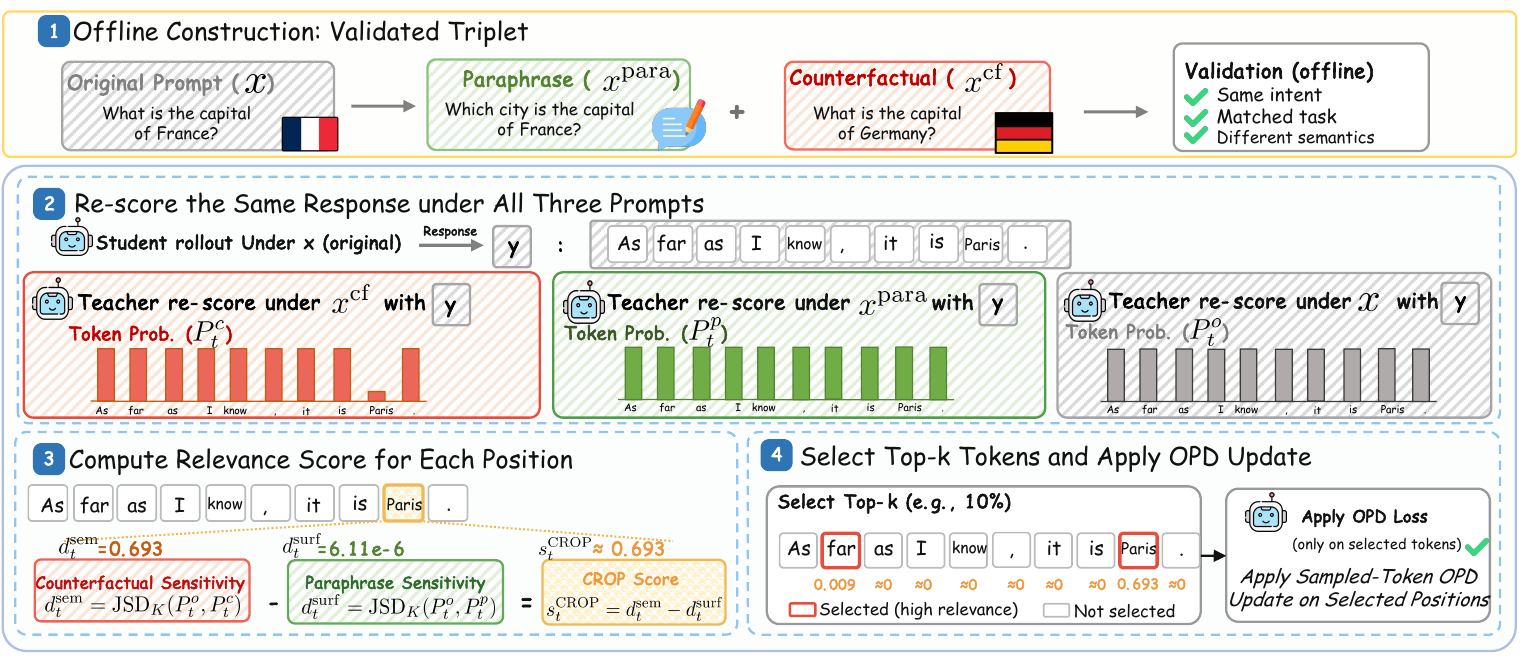}
\caption{Overview of CROP. Validated original-paraphrase-counterfactual triplets enable matched teacher rescoring of a fixed student rollout. CROP ranks tokens by counterfactual minus paraphrase sensitivity and applies a fixed-budget mask independently within each response for sampled-token OPD.}
\label{fig1}
\end{figure*}

Let $\pi_\theta$ denote the trainable student, $\pi_{\bar\theta}$ the
rollout-time student snapshot, and $\pi_T$ the teacher; index $i$ identifies a
rollout sample and $t$ a response-token position. 
For an original prompt $x$, the student produces an on-policy response
$y=(y_1,\ldots,y_T)\sim\pi_{\bar\theta}(\cdot\mid x)$, where $T$ is the
generated response length. 
CROP constructs an offline matched triplet
$(x,x^{\mathrm{para}},x^{\mathrm{cf}})$, where $x^{\mathrm{para}}$ is a
meaning-preserving paraphrase of $x$ and $x^{\mathrm{cf}}$ changes one
task-relevant condition. It measures counterfactual relevance on the fixed
response $y$ and converts the resulting scores into a binary response-token
mask. 
The mask determines which sampled tokens contribute to the existing OPD update. 
The teacher target and per-token OPD objective remain unchanged, while selected
losses are aggregated using the token mean employed by the implementation.
Figure~\ref{fig1} summarizes this pipeline.

\subsection{Counterfactual Triplet Generation}

For each source prompt, we construct an offline original-paraphrase-counterfactual triplet. 
The paraphrase preserves task-relevant meaning, the requested task, and output constraints. 
The counterfactual changes one explicit material condition while preserving the requested task and output constraints~\citep{stolfo2023causal,li2024gsmplus}.
It serves as a validated approximate semantic intervention used to construct an operational token-selection contrast.
An LLM generates candidate rewrites and a separate critic pass validates them
using the prompts in Listings~\ref{lst:generator-system}--\ref{lst:validator-user};
the default training adapter uses only strictly validated triplets.
This validation does not require a unique reference answer: for open-ended tasks, it instead checks the semantic relation between prompts together with task-specific response constraints. 
Although our current triplet-generation prompts are instantiated for mathematics training data, this construction can extend to coding, open-domain question answering, and other tasks for which meaning-preserving paraphrases and condition-changing rewrites can be validated~\citep{jain2021contracode,paranjape2022retrieval}. 
The complete prompts, validation criteria, repair and ranking procedure, and output filtering policy are given in Appendix~\ref{app:counterfactual-generation}.

\subsection{CROP: Paraphrase-calibrated Counterfactual Token Selection}

Given a validated triplet and a student rollout on the original prompt, CROP ranks response positions by whether the teacher distribution shifts more under the condition-changing prompt than under the meaning-preserving paraphrase.
It first uses the teacher to rescore the fixed rollout under the three matched prompts, forms a paraphrase-calibrated counterfactual sensitivity margin, and then allocates a per-response token budget.
Here, the per-response token budget controls how much token-level supervision enters the OPD update instead of the total training computation.
The resulting binary mask changes only which response tokens enter the sampled-token OPD loss; the teacher target and underlying update remain unchanged.

\paragraph{Matched teacher rescoring.}
CROP uses the teacher to evaluate a prompt intervention on a fixed student response. 
Given an original prompt $x$, the rollout-time student produces
$y=(y_1,\ldots,y_T)\sim\pi_{\bar\theta}(\cdot\mid x)$. 
Both paired prompts reuse this original rollout. 
For every response position $t$, the same response token $y_t$ and prefix $y_{<t}$ are scored under the original prompt, a meaning-preserving paraphrase, and a condition-changing counterfactual:
\begin{equation}
P_t^o=\pi_T(\cdot\mid x,y_{<t}),\qquad
P_t^p=\pi_T(\cdot\mid x^{\mathrm{para}},y_{<t}),\qquad
P_t^c=\pi_T(\cdot\mid x^{\mathrm{cf}},y_{<t}).
\label{eq:crop-matched-distributions}
\end{equation}
We omit the rollout-sample index $i$ here and restore it where needed below.
Holding the realized response and all prefixes fixed prevents differences in independently sampled trajectories from being mistaken for prompt sensitivity. 
The three rescoring calls use the fixed teacher and are detached from the subsequent parameter update. 
CROP constructs its score from the teacher, which also supplies the subsequent sampled-token OPD target.

\paragraph{Top-$K$ Jensen-Shannon divergence.}
The teacher scoring endpoint returns the $K$ highest-probability vocabulary
entries at each position, where $K$ is the retained vocabulary-support size. 
For two scored distributions $P$ and $Q$, let
$V_{P,Q}=\operatorname{TopK}(P)\cup\operatorname{TopK}(Q)$. 
We assign zero probability to omitted token IDs and add a residual category for all omitted probability mass:
\begin{equation}
\widetilde R=\bigl(\{R(v)\}_{v\in V_{P,Q}},1-\textstyle\sum_{v\in V_{P,Q}}R(v)\bigr),\qquad R\in\{P,Q\}.
\label{eq:crop-residual-support}
\end{equation}
CROP computes Jensen-Shannon divergence on this finite support,
\begin{equation}
\operatorname{JSD}_K(P,Q)=\frac12D_{\mathrm{KL}}(\widetilde P\Vert M)+
\frac12D_{\mathrm{KL}}(\widetilde Q\Vert M),\qquad
M=\frac12(\widetilde P+\widetilde Q).
\label{eq:crop-jsd}
\end{equation}
where $D_{\mathrm{KL}}$ denotes Kullback--Leibler divergence.
Thus, the implementation score is a top-$K$-with-residual approximation, not an exact full-vocabulary JSD.

\paragraph{Paraphrase-calibrated score.}
For each response position, CROP computes counterfactual and paraphrase sensitivities,
\begin{equation}
d_t^{\mathrm{sem}}=\operatorname{JSD}_K(P_t^o,P_t^c),\qquad
 d_t^{\mathrm{surf}}=\operatorname{JSD}_K(P_t^o,P_t^p),
\end{equation}
and ranks positions using
\begin{equation}
s_t^{\mathrm{CROP}}=d_t^{\mathrm{sem}}-d_t^{\mathrm{surf}}.
\label{eq:crop-score}
\end{equation}
The two terms record distributional change under the condition-changing rewrite and the meaning-preserving paraphrase, respectively.
Their signed difference is used as a paraphrase-calibrated sensitivity margin for ranking; because JSD is nonlinear, it is not an additive decomposition of semantic and surface effects.
A high margin means that the counterfactual induces a larger distribution shift than the paraphrase within the matched triplet.
A low margin can reflect either stability or failure to react to a relevant condition, while a high margin can include overreaction to residual prompt differences.
Accordingly, the score is a model-internal, contrast-specific ranking heuristic, not a measure of correctness, ground-truth task relevance, or an unrestricted causal effect.

\paragraph{Selection under a per-response token budget.}
Let $a_{i,t}\in\{0,1\}$ denote the original response loss mask and define the valid
candidate positions for response $i$ as
\begin{equation}
\mathcal V_i=\{t:a_{i,t}=1,\ s_{i,t}^{\mathrm{CROP}}\in\mathbb R\}.
\end{equation}
Let $\rho\in(0,1]$ be the nominal retention ratio and $L$ the minimum
number of selected positions per nonempty response, with $L=1$ in our
implementation.  Each response receives its own budget
\begin{equation}
b_i=\min\!\left\{
|\mathcal V_i|,
\max\!\left\{1,L,\left\lfloor\rho|\mathcal V_i|\right\rfloor\right\}
\right\}.
\label{eq:crop-response-budget}
\end{equation}
For every response with $|\mathcal V_i|>0$, CROP independently retains
\begin{equation}
\mathcal P_i=\operatorname{TopK}
\left(\mathcal V_i,s_i^{\mathrm{CROP}},b_i\right).
\end{equation}
The final loss mask is
\begin{equation}
m_{i,t}^{\mathrm{CROP}}=
\begin{cases}
a_{i,t}\mathbf{1}\{t\in\mathcal P_i\},&|\mathcal V_i|>0,\\
a_{i,t},&|\mathcal V_i|=0.
\end{cases}
\label{eq:crop-mask}
\end{equation}
Here, $\mathbf{1}\{\cdot\}$ denotes the indicator function.
Thus, tokens compete only with positions from the same response. The floor in
Equation~\ref{eq:crop-response-budget} can make the realized fraction slightly
smaller than $\rho$, whereas minimum-one retention can make it larger for short
responses. A negative score can still be retained when it ranks within that
response's budget.

\paragraph{Masked sampled-token OPD update.}
The underlying OPD implementation requires only sampled log probabilities for
each observed response token. We define the sampled
rollout-student-minus-teacher log-probability gap as
\begin{equation}
d_{i,t}^{\mathrm{OPD}}=\log\pi_{\bar\theta}(y_{i,t}\mid x_i,y_{i,<t})-
\log\pi_T(y_{i,t}\mid x_i,y_{i,<t}).
\end{equation}
Let $A_{i,t}^{\mathrm{base}}$ denote the token-level advantage supplied by the
underlying policy-optimization objective, and let $\lambda_{\mathrm{OPD}}\geq0$
denote the strength of the sampled-token teacher correction. The resulting
token-level OPD advantage is
\begin{equation}
A_{i,t}=A_{i,t}^{\mathrm{base}}-
\lambda_{\mathrm{OPD}}d_{i,t}^{\mathrm{OPD}}.
\end{equation}
The proximal policy optimization (PPO) ratio~\citep{schulman2017ppo} is
\begin{equation}
r_{i,t}(\theta)=\exp\!\left(
\log\pi_\theta(y_{i,t}\mid x_i,y_{i,<t})-
\log\pi_{\bar\theta}(y_{i,t}\mid x_i,y_{i,<t})
\right).
\end{equation}
Let $\epsilon\geq0$ and $\epsilon_{\mathrm{high}}\geq0$ be the lower- and
upper-side clipping radii of the PPO probability ratio, respectively, and let
$\operatorname{clip}(z,l,u)=\min\{\max\{z,l\},u\}$. The underlying OPD update
then uses the unchanged per-token clipped surrogate loss
\begin{equation}
\ell_{i,t}^{\mathrm{sampled\text{-}OPD}}
=\max\!\left[
-r_{i,t}(\theta)A_{i,t},
-\operatorname{clip}\!\left(
r_{i,t}(\theta),1-\epsilon,1+\epsilon_{\mathrm{high}}
\right)A_{i,t}
\right].
\end{equation}
For a global training batch $\mathcal B_{\mathrm{train}}$, CROP uses the
token mean over all selected positions,
\begin{equation}
\mathcal L_{\mathrm{CROP}}=
\frac{
\sum_{i\in\mathcal B_{\mathrm{train}}}\sum_t
m_{i,t}^{\mathrm{CROP}}\ell_{i,t}^{\mathrm{sampled\text{-}OPD}}
}{
\max\!\left\{
\sum_{i\in\mathcal B_{\mathrm{train}}}\sum_t m_{i,t}^{\mathrm{CROP}},1
\right\}}.
\label{eq:crop-loss}
\end{equation}
Thus, CROP is a hard response-token selector. 
It leaves the student rollout distribution, teacher target, and sampled-token OPD signal unchanged. 
Complete rollout generation, OPD teacher scoring, and the three matched teacher rescoring passes still occur before the mask is applied. 
Algorithm~\ref{alg:crop-training} summarizes the complete training procedure.

\begin{algorithm}[t]
\caption{CROP Training}
\label{alg:crop-training}
\begin{algorithmic}[1]
\STATE Construct and validate $(x,x^{\mathrm{para}},x^{\mathrm{cf}})$ offline.
\FOR{each prompt batch $x$}
\STATE Sample $y\sim\pi_{\bar\theta}(\cdot\mid x)$.
\STATE Use $\pi_T$ to rescore fixed $y$ under $x$, $x^{\mathrm{para}}$, and $x^{\mathrm{cf}}$.
\STATE Compute $s_t^{\mathrm{CROP}}$ using Equations~\ref{eq:crop-jsd}--\ref{eq:crop-score}.
\STATE Normalize selector metrics over the rollout batch when required.
\STATE Select the top-$b_i$ positions independently within each response.
\STATE Query $\pi_T$ on original-prompt, student-visited prefixes.
\STATE Update $\pi_\theta$ with the masked sampled-token OPD loss in Equation~\ref{eq:crop-loss}.
\ENDFOR
\end{algorithmic}
\end{algorithm}
\section{Experiments}

We evaluate whether CROP improves sampled-token OPD under a controlled
per-response token budget, whether its ranking is more useful than matched
controls, and how its behavior changes with the nominal retention ratio.

\subsection{Experimental Setup}

\paragraph{Models and training data.}
We study Qwen3-4B-Base-GRPO $\rightarrow$ Qwen3-1.7B-Base (Setting A) and
Qwen3-8B (GRPO) $\rightarrow$ Qwen3-4B-Base (Setting B), covering two teacher-student scales within the same model family for controlled comparison. The latter teacher is
the public Open-R1 GRPO checkpoint~\citep{hdong2026qwen3grpo}. Starting from
17,398 deduplicated DAPO-Math-17K prompts~\citep{yu2026dapo}, triplet
validation yields 16,594 examples shared by all methods
(Appendix~\ref{app:triplet-construction-statistics}). All runs follow the same
115-step training protocol, with 144 prompts and four on-policy
responses per prompt in each complete batch. Exact filtering counts, response
limits, optimization settings, and hardware details are in Appendices
\ref{app:training-hyperparameters} and \ref{app:reproducibility-details}.

\paragraph{Baselines.}
We include the untrained Base model and three dense baselines: Pure OPD~\citep{yang2026learning,song2026survey},
IW-OPD~\citep{xie2026position}, and
our dense adaptation of CREDIT~\citep{shen2026credit}. The hard selectors are
Entropy~\citep{jin2026eopd}, TIP~\citep{xu2026tip}, and TA-OPD~\citep{wang2026taopd};
CROP completes the main comparison. Dense
methods use all valid response positions, whereas every hard selector uses a
20\% nominal retention ratio with the same rule for computing the per-response
token budget in the main results. CF-OPD, PC-OPD,
Random, and CROP-Bottom serve as matched selector controls. This setup allows us to compare CROP against both full-token dense supervision and matched-budget selective baselines. Appendix
\ref{app:baseline-implementation} gives all definitions and adaptation details.

\paragraph{Evaluation.}
We evaluate AIME24~\citep{zhang2024aime}, AIME25~\citep{zhang2025aime},
AMC23~\citep{mathai2025amc23}, and MATH-500~\citep{hendrycks2021math}.
AIME24, AIME25, and AMC23 report Mean@8/Pass@8 from eight samples; the
492-problem MATH-500 subset reports Mean@4/Pass@4 from four samples
(Appendix~\ref{app:contamination-audit}). Avg. is the arithmetic mean of the four
Mean metrics computed from their empirical correct counts before display
formatting. Setting A uses thinking mode, whereas Setting B uses non-thinking
decoding. Appendix~\ref{app:evaluation-protocol} reports the complete decoding
configurations.

\subsection{Main Results}

Table~\ref{tab:main_results} reports the main results. Throughout this paper, we define the {\textit{strongest non-CROP baseline}} as the highest-Avg. method among all non-CROP methods, and the strongest {\textit{non-CROP hard selector}} as the highest-Avg. method among Entropy, TIP, and TA-OPD.

In Setting A, CROP obtains the best Avg. of 26.98, improving over Pure OPD by
2.99 points and over the strongest non-CROP baseline, i.e., TA-OPD, by 1.43 points.
The gains are distributed across multiple benchmarks. CROP has the highest AIME24 Mean, AMC23 Mean, and
MATH-500 Mean and Pass, while tying the best AIME24 and AMC23 Pass rates. Although it does not lead on AIME25, its aggregate advantage is supported by consistent improvements across the remaining benchmarks, not by a single dominant metric.

In Setting B, CROP achieves an Avg. of 52.22, outperforming the strongest non-CROP baseline, i.e., IW-OPD, by 0.96 points and the strongest non-CROP hard selector, i.e., TA-OPD, by 1.44 points. CROP obtains the highest AIME24 Mean and ties the best AIME24 Pass, AIME25 Mean and Pass, and AMC23 Pass rates. On MATH-500, TIP
obtains the highest Mean, while Dense CREDIT obtains the highest Pass rate.

All within-setting rows share the same data and evaluation protocol. Appendix Table~\ref{tab:main-generation-diagnostics} reports available generation
length, format-failure, and truncation diagnostics, providing additional context for the observed performance differences.

\begin{table*}[t]
\centering
\caption{Main results (\%). AIME24/25 and AMC23 report Mean/Pass@8; the
492-problem MATH-500 subset reports Mean/Pass@4. Selectors use a matched 20\%
per-response token budget; dense methods use all valid tokens. Avg. averages
the four unrounded Mean scores. Within-setting maxima, including ties, are
bolded.}
\label{tab:main_results}
\scriptsize
\setlength{\tabcolsep}{2.6pt}
\renewcommand{\arraystretch}{1.02}
\begin{adjustbox}{width=0.93\textwidth,center}
\begin{tabular}{llrrrrrrrrr}
\toprule
Setting & Method & Avg. & \multicolumn{2}{c}{AIME24} & \multicolumn{2}{c}{AIME25} &
\multicolumn{2}{c}{AMC23} & \multicolumn{2}{c}{MATH-500} \\
\cmidrule(lr){4-5}\cmidrule(lr){6-7}\cmidrule(lr){8-9}\cmidrule(lr){10-11}
& & & Mean@8 & Pass@8 & Mean@8 & Pass@8 & Mean@8 & Pass@8 & Mean@4 & Pass@4 \\
\midrule
\multirow{8}{*}{\makecell[l]{Setting A\\4B-Base-GRPO\\$\rightarrow$ 1.7B-Base}}
& Base & 21.15 & 2.92 & 13.33 & 2.08 & 6.67 & 26.56 & 62.50 & 53.05 & 73.37 \\
& Pure OPD & 23.99 & 4.17 & 13.33 & 3.33 & 13.33 & 30.63 & 62.50 & 57.83 & 77.03 \\
& Entropy & 23.72 & 4.17 & \textbf{23.33} & 3.33 & 13.33 & 29.06 & 60.00 & 58.33 & 78.25 \\
& TIP & 23.95 & 4.58 & 20.00 & 2.92 & 10.00 & 30.94 & 65.00 & 57.37 & 76.42 \\
& Dense CREDIT & 25.27 & 4.17 & 20.00 & \textbf{4.17} & \textbf{16.67} & 34.06 & 65.00 & 58.69 & 75.20 \\
& IW-OPD & 24.65 & 3.75 & 10.00 & 3.75 & 13.33 & 33.13 & 65.00 & 57.98 & 76.02 \\
& TA-OPD & 25.55 & 5.42 & 20.00 & 2.92 & 10.00 & 33.75 & \textbf{67.50} & 60.11 & 78.66 \\
& \cellcolor{blue!18}\textbf{CROP}
& \cellcolor{blue!18}\textbf{26.98}
& \cellcolor{blue!18}\textbf{8.33} & \cellcolor{blue!18}\textbf{23.33}
& \cellcolor{blue!18}3.75 & \cellcolor{blue!18}10.00
& \cellcolor{blue!18}\textbf{35.00} & \cellcolor{blue!18}\textbf{67.50}
& \cellcolor{blue!18}\textbf{60.82} & \cellcolor{blue!18}\textbf{78.86} \\
\midrule
\multirow{8}{*}{\makecell[l]{Setting B\\8B-GRPO\\$\rightarrow$ 4B-Base}}
& Base & 48.65 & 24.17 & 46.67 & \textbf{24.58} & 33.33 & 63.75 & 85.00 & 82.11 & 91.67 \\
& Pure OPD & 48.94 & 22.92 & 43.33 & 20.00 & 30.00 & 69.38 & 90.00 & 83.49 & 89.63 \\
& Entropy & 50.75 & 24.17 & 36.67 & 22.08 & 36.67 & 71.88 & 85.00 & 84.86 & 90.65 \\
& TIP & 50.35 & 24.17 & 40.00 & 21.67 & 40.00 & 70.63 & \textbf{92.50} & \textbf{84.96} & 92.48 \\
& Dense CREDIT & 50.22 & 22.08 & 40.00 & 21.67 & 33.33 & \textbf{75.00} & 85.00 & 82.11 & \textbf{93.09} \\
& IW-OPD & 51.26 & 27.50 & \textbf{50.00} & 22.50 & 43.33 & 71.25 & \textbf{92.50} & 83.79 & 91.67 \\
& TA-OPD & 50.78 & 23.75 & 40.00 & \textbf{24.58} & \textbf{50.00} & 71.25 & \textbf{92.50} & 83.54 & 92.07 \\
& \cellcolor{blue!18}CROP
& \cellcolor{blue!18}\textbf{52.22}
& \cellcolor{blue!18}\textbf{27.92} & \cellcolor{blue!18}\textbf{50.00}
& \cellcolor{blue!18}\textbf{24.58} & \cellcolor{blue!18}\textbf{50.00}
& \cellcolor{blue!18}72.19 & \cellcolor{blue!18}\textbf{92.50}
& \cellcolor{blue!18}84.20 & \cellcolor{blue!18}92.48 \\
\bottomrule
\end{tabular}
\end{adjustbox}
\end{table*}

\subsection{Selector Ablation}
\label{sec:selector-ablation}
Table~\ref{tab:selector-ablation} separates two design questions at a matched
20\% per-response token budget: whether the paraphrase-calibrated CROP ranking is
useful, and whether the task relevance score is better derived from the student or teacher.
CF-OPD retains only the teacher-scored counterfactual term, whereas PC-OPD
retains only the negative teacher-scored paraphrase term. Random controls for
token sparsity alone, and CROP-Bottom reverses the full teacher-scored CROP
ranking. CROP-student instead computes the full calibrated score from student
distributions. Appendix~\ref{app:crop-student} gives the complete CROP-student
formulation and implementation details.
\begin{table*}[t]
\centering
\caption{Selector ablation in Setting A at a matched 20\% per-response token budget
(\%). CROP-Bottom reverses the teacher-scored CROP ranking, whereas
CROP-student replaces teacher rescoring with student rescoring.
$\Delta=\mathrm{Avg.}_{\mathrm{method}}-\mathrm{Avg.}_{\mathrm{CROP}}$;
column maxima are bolded.}
\label{tab:selector-ablation}
\footnotesize
\setlength{\tabcolsep}{4.5pt}
\renewcommand{\arraystretch}{1.0}
\begin{tabular}{lcccccc}
\toprule
Method & AIME24 & AIME25 & AMC23 & MATH-500 & Avg. & $\Delta$ \\
\midrule
CF-OPD & 4.58 & 4.17 & 29.38 & 59.71 & 24.46 & $-2.52$ \\
PC-OPD & 4.58 & 2.92 & 29.69 & 57.57 & 23.69 & $-3.29$ \\
Random & 5.00 & \textbf{4.58} & 29.38 & 58.74 & 24.42 & $-2.55$ \\
CROP-Bottom & 5.83 & 1.67 & 33.44 & 58.74 & 24.92 & $-2.06$ \\
CROP-student & 5.42 & 4.17 & \textbf{35.62} & 60.26 & 26.37 & $-0.61$ \\
\cellcolor{blue!18}\textbf{CROP}
& \cellcolor{blue!18}\textbf{8.33}
& \cellcolor{blue!18}3.75
& \cellcolor{blue!18}35.00
& \cellcolor{blue!18}\textbf{60.82}
& \cellcolor{blue!18}\textbf{26.98}
& \cellcolor{blue!18}\textbf{0.00} \\
\bottomrule
\end{tabular}
\end{table*}
With teacher rescoring, CROP reaches 26.98 Avg., exceeding CF-OPD by
2.52 points and PC-OPD by 3.29 points. Thus, combining the counterfactual and
paraphrase components into the calibrated CROP score outperforms either
teacher-scored component alone. CROP also exceeds Random by 2.55 points and
CROP-Bottom by 2.06 points, showing that its gains depend on informative token ranking beyond sparse selection alone.
Replacing teacher rescoring with student rescoring (CROP-student) lowers Avg.
from 26.98 to 26.37, suggesting that teacher distributions provide a modestly
stronger aggregate ranking signal in Setting A. This advantage holds across all
tested budgets from 5\% to 20\% (Appendix
Table~\ref{tab:teacher-student-budget}).

\subsection{Selector Behavior across Entropy Ranks}

\begin{figure*}[t]
\centering
\includegraphics[width=0.98\textwidth]{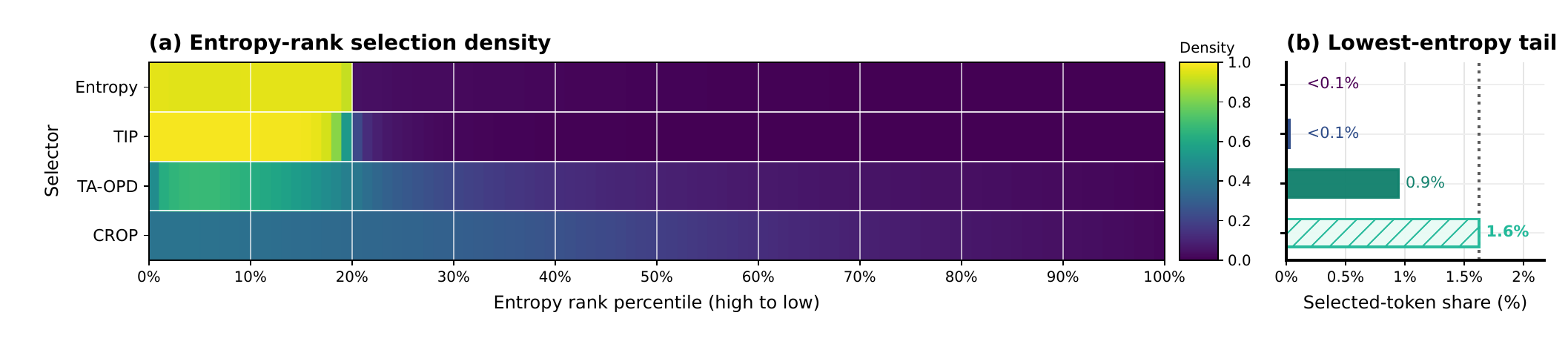}
\caption{Token-selection behavior across student-entropy ranks in Setting A at
the matched 20\% per-response token budget. Panel (a) shows row-normalized selection
density from high to low entropy. Panel (b) reports the selected-token share in
the plotted lowest-entropy tail; the dotted line denotes the
entropy-independent reference.}
\label{fig:selector-entropy-ranks}
\end{figure*}

To examine whether CROP captures a signal distinct from uncertainty,
Figure~\ref{fig:selector-entropy-ranks} compares where different selectors
allocate their retained tokens along the student-entropy ordering in Setting A
under the same 20\% per-response token budget. Entropy and TIP assign less than
0.1\% of their selected tokens to the plotted lowest-entropy tail, whereas CROP
assigns 1.6\%, matching the entropy-independent reference and exceeding TA-OPD
(0.9\%). This behavior indicates that CROP does not simply reproduce an
uncertainty-based ranking, consistent with its premise that low-entropy tokens
can remain sensitive to task-specific conditions. Together with the matched
controls in Section~\ref{sec:selector-ablation}, this supports interpreting CROP
as a relevance-oriented selection signal. Appendix~\ref{app:crop-ent} further
examines its interaction with entropy.

\begin{figure*}[t]
\centering
\includegraphics[width=0.90\textwidth]{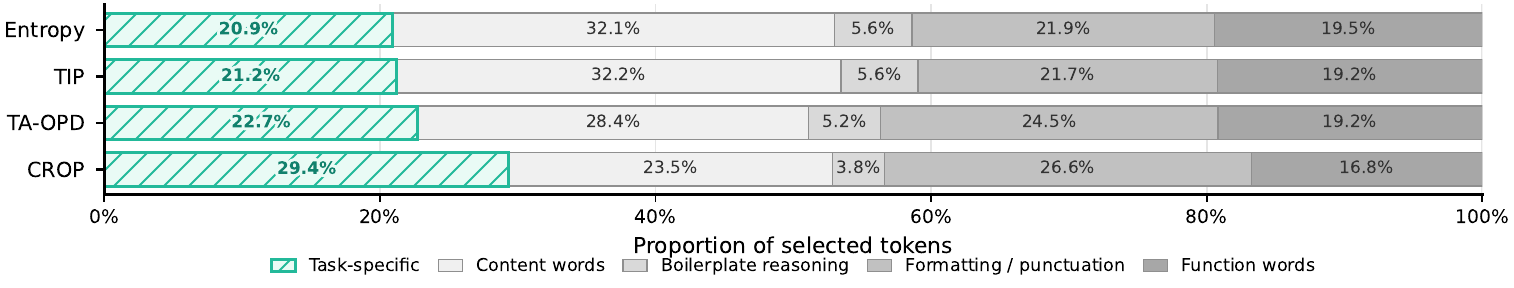}
\caption{Selected-token composition in Setting A at the matched 20\%
per-response budget. Categories are mutually exclusive; task-specific tokens
are highlighted. Values may not total 100\% because of rounding.}
\label{fig:selected-token-categories}
\end{figure*}

To assess whether CROP captures task relevance rather than merely optimization
need, Figure~\ref{fig:selected-token-categories} compares selected-token
composition in Setting A at the matched 20\% budget. Task-specific tokens make
up 29.4\% of CROP's selections, versus 20.9\% for Entropy, 21.2\% for TIP,
and 22.7\% for TA-OPD. This 6.7-point lead provides complementary corpus-level
evidence that CROP prioritizes task-specific supervision.
Appendix~\ref{app:selected-token-categories} details the taxonomy and
classification procedure. Appendix~\ref{app:crop-qualitative-analysis}
presents four controlled cases of localized task-sensitive CROP responses.

\subsection{Budget Sensitivity}

\begin{figure*}[t]
\centering
\includegraphics[width=0.90\textwidth]{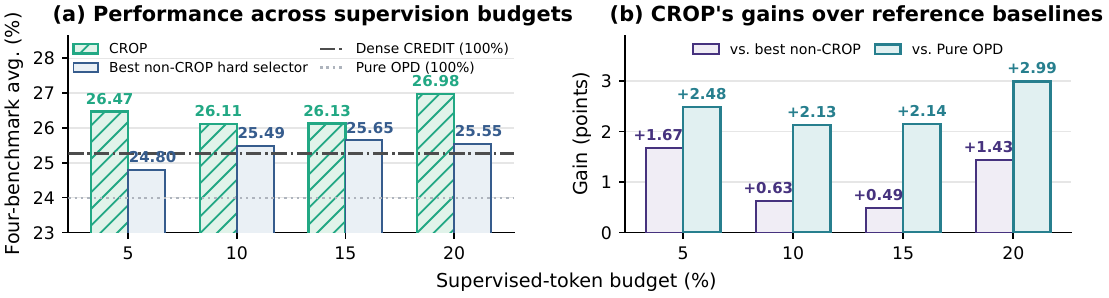}
\caption{Budget sensitivity in Setting A. (a) CROP versus the strongest
matched-budget non-CROP hard selector, with 100\%-supervision Dense CREDIT and
Pure OPD references. (b) CROP gains over the strongest matched-budget non-CROP
hard selector and Pure OPD. Values are four-benchmark macro averages.}
\label{fig:budget-sensitivity}
\end{figure*}

Figure~\ref{fig:budget-sensitivity} tests whether CROP's advantage persists as
the nominal retention ratio changes in Setting A. At each budget, the
hard selector used as the reference is the non-CROP method with the highest Avg. at that
same budget; its identity may therefore change across the sweep. Pure OPD and
Dense CREDIT provide fixed dense references using all valid response tokens.

Across budgets from 5\% to 20\%, CROP achieves average scores of 26.47, 26.11,
26.13, and 26.98, respectively. It outperforms the strongest non-CROP hard
selector by 1.67, 0.63, 0.49, and 1.43 points, so the advantage remains
positive throughout the displayed range but varies substantially in magnitude.
CROP also stays 2.13--2.99 points above Pure OPD and exceeds Dense CREDIT at
every tested budget.
Notably, CROP's performance curve is non-monotonic: the 5\% setting is only 0.51 points below the best 20\% result and outperforms both intermediate budgets. These results highlight CROP’s effectiveness under sparse supervision.
Substantial gains persist even when only 5\% of response tokens are retained, underscoring the importance of token-ranking quality when the supervision budget is limited.

\section{Conclusion}

We introduced CROP, a selective OPD method that operationalizes task relevance
through matched paraphrase and counterfactual interventions. CROP holds the
student rollout fixed, measures how the teacher's next-token distribution changes across
the matched contexts, and retains the highest-ranked positions under the
resulting paraphrase-calibrated counterfactual sensitivity margin. With a 20\%
per-response token budget, CROP achieves
the best aggregate performance. Matched ablations further show that combining counterfactual and paraphrase signals outperforms either component alone, while entropy-rank analysis reveals selection behavior distinct from uncertainty-based methods. Notably, with 5\% of response tokens retained, CROP remains 0.51 points below its best 20\% result and outperforms the strongest matched-budget hard selector by 1.67 points. Overall, the evidence supports task relevance as an important principle for selective distillation and demonstrates that matched semantic interventions can help translate this principle into token-level supervision.

\bibliographystyle{iclr2027_conference}
\bibliography{references}

\appendix

\section{Counterfactual Triplet Construction and Validation}
\label{app:counterfactual-generation}

\subsection{Prompt Templates}

We generate complete problem rewrites directly. Listings~\ref{lst:generator-system}--\ref{lst:validator-user} reproduce the generator and critic templates used by the data-generation script. The placeholders \texttt{\{problem\}}, \texttt{\{candidate\_count\}}, \texttt{\{feedback\}}, \texttt{\{original\}}, and \texttt{\{candidates\_json\}} are populated for each request. Generation and validation use GPT-5.1 in separate API calls with different prompts; the critic is therefore a separate validation pass, not a distinct model. Human inspections are also incorporated as an independent check on semantic preservation and counterfactual validity.

\begin{lstlisting}[caption={Generator system prompt.},label={lst:generator-system},basicstyle=\scriptsize\ttfamily]
You generate matched mathematical problem interventions.

Given one original math problem, write:
1. a complete paraphrase problem that preserves every mathematical condition,
   task objective, answer type, and correct answer;
2. a complete counterfactual problem that changes exactly one explicit,
   material mathematical condition while preserving the task objective and
   answer type.

You may rewrite the full wording naturally. You do not need to identify or copy
an exact substring. Do not solve the problem, reveal an answer, add hidden
assumptions, or change multiple conditions. The counterfactual must remain
coherent and should have a unique finite answer. Return JSON only.
\end{lstlisting}

\begin{lstlisting}[caption={Generator user prompt.},label={lst:generator-user},basicstyle=\scriptsize\ttfamily]
Generate up to {candidate_count} distinct matched intervention candidates.

Original problem:
<<<
{problem}
>>>

Requirements:
- paraphrase_problem must be a complete standalone rewrite of the original
  problem and must preserve its exact mathematical meaning;
- counterfactual_problem must be a complete standalone problem that changes
  exactly one explicit material condition;
- preserve the requested task, output format, object identities, and answer
  type;
- do not include a solution, final answer, commentary, Markdown fences, or the
  wrapper instructions outside the math problem;
- prefer different changed conditions across candidates;
- both generated problems must be grammatical and valid mathematical text.

{feedback}

Return exactly:
{
  "decision": "candidates",
  "candidates": [
    {
      "condition_type": "short category",
      "original_condition_description": "the original explicit condition",
      "paraphrased_condition_description": "the equivalent condition wording",
      "counterfactual_condition_description": "the changed condition",
      "paraphrase_problem": "complete paraphrased math problem",
      "counterfactual_problem": "complete counterfactual math problem",
      "counterfactual_description": "what changed and why it is one condition"
    }
  ]
}
\end{lstlisting}

\begin{lstlisting}[caption={Critic system prompt.},label={lst:validator-system},basicstyle=\scriptsize\ttfamily]
You are a conservative independent critic of mathematical dataset interventions.
Judge the supplied complete rewrites rather than comparing exact character spans.
Reject when uncertain.

Check whether the paraphrase preserves all mathematical meaning and whether the
counterfactual changes exactly one explicit condition, introduces no unintended
assumption, remains coherent and determinate, and preserves the original task
and answer type. Give separate confidence values and concrete reasons for the
paraphrase and counterfactual. Evaluate every candidate independently before
ranking the accepted candidates; do not let one candidate change the standard
applied to another. Return JSON only.
\end{lstlisting}

\begin{lstlisting}[caption={Critic user prompt.},label={lst:validator-user},basicstyle=\scriptsize\ttfamily]
Evaluate all generated candidates for the same original problem.

Original problem:
<<<
{original}
>>>

Candidates, numbered from 1:
{candidates_json}

Return exactly:
{
  "evaluations": [
    {
      "candidate_number": 1,
      "decision": "<accept|reject>",
      "violations": [{"criterion": "short_snake_case_name",
                      "evidence": "specific evidence from the triplet"}],
      "paraphrase_status": "<accept|reject>",
      "paraphrase_confidence": 0.0,
      "paraphrase_reason": "specific equivalence judgment",
      "counterfactual_status": "<accept|reject>",
      "counterfactual_confidence": 0.0,
      "counterfactual_reason": "specific one-condition, coherence, and determinacy judgment",
      "original_answer_type": "one allowed answer type",
      "counterfactual_answer_type": "one allowed answer type",
      "original_determinacy": "<determinate|underdetermined|inconsistent|unknown>",
      "counterfactual_determinacy": "<determinate|underdetermined|inconsistent|unknown>",
      "determinacy_evidence": "concrete mathematical evidence",
      "confidence": 0.0
    }
  ],
  "selected_candidate_number": 1
}

Allowed answer types: single_integer, single_rational, single_real, finite_set,
expression, proof, yes_no, unknown.

For each candidate, use decision="accept" only if both individual statuses are
accept, violations is empty, both problems are determinate, and the answer type
is preserved. Set selected_candidate_number to the highest-confidence accepted
candidate, or null when none is accepted. Evaluate every supplied candidate.
\end{lstlisting}

\subsection{Generation, Validation, and Filtering}

The generator produces up to three candidates in each round and can run two rounds by default. Failure reasons from the critic are supplied as feedback for a subsequent repair round. The separate critic call evaluates every candidate. A candidate passes only if its overall decision and both individual statuses are \texttt{accept}, the violation list is empty, all three confidence values meet the configured threshold (0.85 by default), the answer types agree, and both problems are determinate. Passing candidates are ranked by overall confidence, then by the smaller of the paraphrase and counterfactual confidence, followed by generation round and candidate index.

Records are labelled \texttt{strict\_pass} when they pass in the first round, \texttt{repaired\_pass} when they pass later, \texttt{needs\_review} when a complete triplet remains below the strict criterion, and \texttt{pending\_repair} when no complete triplet is produced. The generation script writes every complete triplet to its training output. The default CROP dataset adapter retains records with \texttt{triplet\_complete}, \texttt{usable\_for\_training}, and \texttt{validation\_passed} all true. Thus the CROP training runs use the validated subset.

\subsection{Aggregate Construction Statistics}
\label{app:triplet-construction-statistics}

Table~\ref{tab:triplet-generation-funnel} summarizes the terminal outcome for
all 17,398 source prompts. Of these, 16,148 passed in the first generation
round. The remaining 607 prompts that reached generation entered the
feedback-based repair round, and 446 subsequently passed, for a repair success
rate of 73.48\% within this repair-eligible subset. The final validated training
subset therefore contains 16,594 triplets, or 95.38\% of all source prompts.
The 643 pre-generation exclusions were caused by control characters in the
source problem and are pipeline-input failures, not semantic rejections
of generated triplets. Among the 18 generation-stage incomplete records, 17
contained control characters in a generated rewrite and one ended after an API
timeout.

\begin{table}[t]
\centering
\caption{Terminal outcomes of the triplet-construction pipeline. Percentages
use all 17,398 source prompts as the denominator. The final validated subset is
the sum of strict and repaired passes.}
\label{tab:triplet-generation-funnel}
\small
\setlength{\tabcolsep}{5pt}
\begin{tabular}{@{}p{0.58\columnwidth}rr@{}}
\toprule
Terminal outcome & Count & Input share \\
\midrule
First-round strict pass & 16,148 & 92.82\% \\
Repaired pass & 446 & 2.56\% \\
Complete but needs review & 143 & 0.82\% \\
Incomplete after generation/repair & 18 & 0.10\% \\
Pre-generation input exclusion & 643 & 3.70\% \\
\midrule
Final validated training subset & 16,594 & 95.38\% \\
\bottomrule
\end{tabular}
\end{table}

For the accepted records, all three critic-assigned confidence fields are
present. Table~\ref{tab:triplet-confidence} reports them as descriptive
diagnostics only. These values are self-reported by the critic model and are
not calibrated estimates of corpus-level correctness.

\begin{table}[t]
\centering
\caption{Critic-assigned confidence over the 16,594 accepted triplets.}
\label{tab:triplet-confidence}
\small
\setlength{\tabcolsep}{4.5pt}
\begin{tabular}{lrrrr}
\toprule
Decision & Mean & 5th pct. & Median & Minimum \\
\midrule
Paraphrase & 0.975 & 0.95 & 0.98 & 0.86 \\
Counterfactual & 0.944 & 0.90 & 0.95 & 0.85 \\
Overall & 0.951 & 0.90 & 0.95 & 0.85 \\
\bottomrule
\end{tabular}
\end{table}

\subsection{Intervention-Type Distribution}
\label{app:intervention-type-distribution}

The generator records a short free-form \texttt{condition\_type} instead of
selecting from a predefined ontology. All 16,594 accepted records contain this
field, but the corpus has 9,116 distinct exact strings after lowercasing. To
avoid imposing an unverifiable post-hoc semantic taxonomy,
Table~\ref{tab:condition-type-distribution} reports the most frequent exact
labels. The long tail reflects lexical fragmentation as well as intervention
diversity; for example, \texttt{side length}, \texttt{segment length}, and
\texttt{circle radius} are retained as distinct generator labels.

\begin{table}[t]
\centering
\caption{Most frequent exact condition-type labels among the 16,594 accepted
triplets. Labels are lowercased but otherwise not semantically merged.}
\label{tab:condition-type-distribution}
\small
\setlength{\tabcolsep}{5pt}
\begin{tabular}{lrr}
\toprule
Exact condition-type label & Count & Share \\
\midrule
side length & 478 & 2.88\% \\
equation constant & 187 & 1.13\% \\
modulus & 123 & 0.74\% \\
segment length & 115 & 0.69\% \\
range of n & 87 & 0.52\% \\
initial value & 75 & 0.45\% \\
sum value & 74 & 0.45\% \\
circle radius & 74 & 0.45\% \\
summation upper limit & 68 & 0.41\% \\
angle measure & 68 & 0.41\% \\
grid size & 67 & 0.40\% \\
domain of variables & 67 & 0.40\% \\
All other exact labels & 15,111 & 91.06\% \\
\bottomrule
\end{tabular}
\end{table}

\subsection{Final Rejection Diagnostics}
\label{app:triplet-rejection-diagnostics}

Table~\ref{tab:triplet-rejection-diagnostics} summarizes structured diagnostic
fields for the 143 complete triplets whose final status is
\texttt{needs\_review}. The rows are non-exclusive: a record can, for example,
have both a rejected counterfactual and an answer-type mismatch. The counts
refer to the final retained fallback candidate for each source problem rather
than to every rejected candidate generated in earlier rounds. Twenty records
have no structured final critic result because the critic request or returned
schema failed; they remain excluded from training.

\begin{table}[t]
\centering
\caption{Non-exclusive final diagnostics among the 143 complete
\texttt{needs\_review} records.}
\label{tab:triplet-rejection-diagnostics}
\small
\setlength{\tabcolsep}{5pt}
\begin{tabular}{@{}p{0.58\columnwidth}rr@{}}
\toprule
Diagnostic & Count & Share \\
\midrule
Counterfactual rejected & 77 & 53.85\% \\
Counterfactual non-determinate & 52 & 36.36\% \\
Paraphrase rejected & 27 & 18.88\% \\
Counterfactual confidence below 0.85 & 23 & 16.08\% \\
Overall confidence below 0.85 & 22 & 15.38\% \\
Answer-type mismatch & 21 & 14.69\% \\
Missing structured critic result & 20 & 13.99\% \\
Original problem non-determinate & 19 & 13.29\% \\
Paraphrase confidence below 0.85 & 12 & 8.39\% \\
\bottomrule
\end{tabular}
\end{table}

Together with the aggregate diagnostics above, the examples below provide
qualitative illustrations of accepted and rejected interventions. They are not
used to estimate corpus-wide accuracy.

\subsection{Qualitative Accepted and Rejected Examples}
\label{app:counterfactual-examples}

The following triplets illustrate the distinction between a valid local
intervention and a superficially local rewrite that breaks mathematical
coherence. We remove only the common answer-format wrapper for readability. The
accepted example is a first-round \texttt{strict\_pass} record used by CROP; the
rejected example is a \texttt{needs\_review} record excluded from the final
16,594-example training subset. In the examples below,
\sourcecondition{blue} marks the original condition,
\acceptedcondition{green} marks an accepted counterfactual replacement, and
\rejectedcondition{red} marks a rejected replacement. Ordinary paraphrasing
differences are left uncolored.

\paragraph{Accepted example (\texttt{dapo\_math\_003119}).}
\begin{description}
    \item[Original.] Mr.~J left his entire estate to his wife, his daughter, his
    son, and the cook. His daughter and son got half the estate, sharing in the
    ratio $4:3$. His wife got twice as much as the son. If the cook received a
    bequest of \sourcecondition{\$500}, what was the total value of the estate?

    \item[Paraphrase.] Mr.~J left all of his estate to four individuals: his
    wife, his daughter, his son, and the cook. The daughter and son together
    were allotted one-half of the estate, and their shares were divided in the
    ratio $4:3$. His wife's portion was equal to twice the amount given to the
    son. The cook's inheritance amounted to \sourcecondition{\$500}. What was the total value of
    Mr.~J's estate?

    \item[Counterfactual.] Mr.~J left all of his estate to four individuals:
    his wife, his daughter, his son, and the cook. The daughter and son together
    were allotted one-half of the estate, and their shares were divided in the
    ratio $4:3$. His wife's portion was equal to twice the amount given to the
    son. The cook's inheritance amounted to \acceptedcondition{\$600}. What was the total value of
    Mr.~J's estate?
\end{description}

\noindent\textit{Why it is accepted.}
Let the total estate be $E$. The son receives
$\frac{3}{7}\cdot\frac{E}{2}=\frac{3E}{14}$, so the wife receives
$\frac{3E}{7}$. Together with the daughter-son half, the three family members
receive
\begin{equation}
\frac{E}{2}+\frac{3E}{7}=\frac{13E}{14},
\end{equation}
leaving $E/14$ for the cook. The original and counterfactual answers are
$14\cdot500=7000$ and $14\cdot600=8400$, respectively. Both are unique,
nonzero integers. The rewrite therefore changes exactly one explicit numerical
condition while preserving the task, answer type, and determinacy. The
separate critic pass assigned $0.99$ confidence to the paraphrase,
counterfactual, and overall decision and found no violations.

\paragraph{Accepted non-numerical example (\texttt{dapo\_math\_005780}).}
\begin{description}
    \item[Original.] One fair die has faces $1,1,2,2,3,3$, and another has
    faces $4,4,5,5,6,6$. The dice are rolled and the numbers on the top faces
    are added. Find the probability that the sum will be
    \sourcecondition{odd}. Write the probability as $k/m$ in lowest terms and
    give $k+m$.

    \item[Paraphrase.] You roll two fair six-sided dice. The first die has faces
    numbered $1,1,2,2,3,3$, and the second has faces numbered
    $4,4,5,5,6,6$. Add the two top-face numbers and determine the probability
    that the resulting sum is \sourcecondition{odd}. Express the answer as a
    reduced fraction $k/m$ and find $k+m$.

    \item[Counterfactual.] You roll the same two fair dice and add the two
    top-face numbers. Determine the probability that the resulting sum is
    \acceptedcondition{even}. Express the answer as a reduced fraction $k/m$
    and find $k+m$.
\end{description}

\noindent\textit{Why it is accepted.}
The first die is odd with probability $2/3$ and even with probability $1/3$;
the second is even with probability $2/3$ and odd with probability $1/3$.
Hence the original odd-sum probability and counterfactual even-sum probability
are, respectively,
\begin{equation}
\Pr(\text{odd})=\frac{2}{3}\frac{2}{3}+\frac{1}{3}\frac{1}{3}
=\frac{5}{9},\qquad
\Pr(\text{even})=\frac{2}{3}\frac{1}{3}+\frac{1}{3}\frac{2}{3}
=\frac{4}{9}.
\end{equation}
The requested outputs are therefore $14$ and $13$, both unique and nonzero.
The intervention changes a categorical event, \emph{odd} to \emph{even}, rather
than replacing a number. All other conditions and the answer type are
preserved. The separate critic pass accepted all three decisions with confidence
$0.99$ and found no violations.

\paragraph{Rejected example (\texttt{dapo\_math\_002535}).}
\begin{description}
    \item[Original.] The \sourcecondition{arithmetic mean} of the nine numbers in
    $\{9,\allowbreak 99,\allowbreak 999,\allowbreak 9999,\allowbreak
    \ldots,\allowbreak 999999999\}$ is a nine-digit number $M$ whose
    digits are all distinct. Which digit is not contained in $M$?

    \item[Paraphrase.] Let $S$ be the set of nine integers
    $\{9,\allowbreak 99,\allowbreak 999,\allowbreak 9999,\allowbreak
    \ldots,\allowbreak 999999999\}$, where each element is formed by
    repeating the digit $9$ some number of times, from one digit up to nine
    digits. Let $M$ be the \sourcecondition{arithmetic mean (average)} of the nine numbers in
    $S$. The value of $M$ is a nine-digit integer whose digits are all different
    from one another. Which digit from $0$ to $9$ does not occur as a digit of
    $M$?

    \item[Counterfactual.] Let $S$ be the set of nine integers
    $\{9,\allowbreak 99,\allowbreak 999,\allowbreak 9999,\allowbreak
    \ldots,\allowbreak 999999999\}$, where each element is formed by
    repeating the digit $9$ some number of times, from one digit up to nine
    digits. Let $M$ be the \rejectedcondition{median} of the nine numbers in $S$. The value of $M$
    is a nine-digit integer whose digits are all different from one another.
    Which digit from $0$ to $9$ does not occur as a digit of $M$?
\end{description}

\noindent\textit{Why it is rejected.}
The median of the ordered nine-element set is its fifth member, $99999$.
Consequently, this local textual edit produces a value with five equal digits,
contradicting the
retained assertion that $M$ is a nine-digit integer with distinct digits. Thus
the counterfactual changes the record from \texttt{determinate} to
\texttt{inconsistent}. The critic rejected it with confidence $0.99$, and the
default adapter excludes it because \texttt{validation\_passed=false}. This
example shows why a one-span edit alone is insufficient: a valid intervention
must also preserve internal consistency, determinacy, task, and answer type.

\section{Implementation Details}
\label{app:baseline-implementation}

\subsection{Shared Selection Protocol}

The hard selectors include Entropy, TIP, TA-OPD, CF-OPD, PC-OPD,
Random, CROP-Bottom, CROP-student, CROP, and CROP-ent. They retain the same
sampled-token OPD update and use method-specific scores to construct the
response loss mask. Within each
nonempty response $i$, a selector ranks finite scores and retains the top
$b_i$ positions from Equation~\ref{eq:crop-response-budget}. Selection is
performed independently within each response. The resulting mask is
binary and the selected losses are aggregated using the token mean in
Equation~\ref{eq:crop-loss}. Batch-level quantile normalization, when used,
rescales scores while preserving this per-response selection scope. Pure OPD,
IW-OPD, and Dense CREDIT are dense methods that retain all valid response
positions.

\paragraph{Realized Retention under the Main Budget}
\label{app:realized-retention}

For a hard selector, the realized retention ratio is the total number of retained
positions divided by the total number of originally valid response positions.
Because Equation~\ref{eq:crop-response-budget} uses a floor together with
minimum-one retention, the realized retention ratio can lie slightly below or above the
nominal ratio.

\subsection{Selector Score Construction}

Entropy and TIP use full-vocabulary student entropy. TIP's disagreement
proxy and TA-OPD use top-$K$ quantities with $K=16$; this vocabulary-support
size is distinct from the per-response token budget.

For a distribution $P$, let $H(P)$ denote its Shannon entropy. CROP-ent uses
student top-$K$ probabilities renormalized over their retained support, with
entropy divided by $\log K$. Metrics are normalized independently over valid
response positions in the completed rollout batch.
For a generic metric $z_{i,t}$, let $z_{\mathcal B}$ collect its values over
valid positions in rollout batch $\mathcal B$, and let
$Q_{\tau}(z_{\mathcal B})$ denote their empirical $\tau$-quantile. Except for
TIP's normalization specified below, we use 5--95\% quantile clipping:
\begin{equation}
\operatorname{Norm}_{\mathcal B}(z_{i,t}) =
\operatorname{clip}\!\left(
\frac{
z_{i,t}-Q_{0.05}(z_{\mathcal B})
}{
Q_{0.95}(z_{\mathcal B})-Q_{0.05}(z_{\mathcal B})
},
0,1
\right),
\label{eq:selector-batch-normalization}
\end{equation}
with zero output when the quantile span is below $10^{-12}$. We denote the
resulting normalized student entropy,
teacher-to-student divergence, and teacher-student compatibility by
$\widetilde H$, $\widetilde D$, and $\widetilde C_{\mathrm{mass}}$,
respectively.

\subsection{Dense OPD Baselines}
\label{app:credit-adaptation}

\paragraph{Pure OPD.}
Pure OPD retains the original response loss mask and applies OPD supervision
to every valid response position.

\paragraph{IW-OPD.}
IW-OPD applies dense continuous weighting to all valid response positions. Let
the sampled absolute log-probability discrepancy be
\begin{equation}
d_{i,t}=\left|
\log\pi_T(y_{i,t}\mid x_i,y_{i,<t})-
\log\pi_{\bar\theta}(y_{i,t}\mid x_i,y_{i,<t})
\right|.
\end{equation}
Let $T_i$ denote the response length for sample $i$, and let
$\varepsilon_{\mathrm{IW}}>0$ be a numerical stabilizer. Define the discrepancy
mass before position $t$ by
\begin{equation}
F_{i,t-1}=\frac{\sum_{k<t}d_{i,k}}
{\sum_{k\leq T_i}d_{i,k}+\varepsilon_{\mathrm{IW}}}.
\end{equation}
Let $w_{\max}$ denote the maximum multiplier assigned at the beginning of a
response. The detached position weight is
\begin{equation}
w_{i,t}=1+(w_{\max}-1)(1-F_{i,t-1}).
\end{equation}
It multiplies the OPD advantage at every valid response position. We use the
absolute-discrepancy variant with $w_{\max}=1.5$, and all valid positions remain
active.

\paragraph{Dense CREDIT.}
Dense CREDIT holds the sampled response fixed and rescores it with the
teacher under unrelated prompts drawn from distinct prompt groups in the same
completed rollout batch. Let $x_{i,j}^{-}$ denote the $j$th unrelated negative
prompt paired with response $i$, let $C$ be the number of negative prompts, and
let $\lambda\geq0$ weight their average contribution. Its contrastive teacher
signal is
\begin{equation}
\widetilde\ell^{T}_{i,t}=
\log\pi_T(y_{i,t}\mid x_i,y_{i,<t})-
\lambda\frac{1}{C}\sum_{j=1}^{C}
\log\pi_T(y_{i,t}\mid x_{i,j}^{-},y_{i,<t}).
\end{equation}
This signal replaces the positive-prompt teacher log probability in the OPD
correction and is applied densely at every valid response position. We use
$C=1$ and $\lambda=0.1$, with negative-selection seed 42. Our fixed external
teacher yields an external-teacher adaptation of CREDIT's EMA
self-distillation formulation.

\subsection{Optimization-oriented Hard Selectors}

\paragraph{Entropy.}
The Entropy baseline ranks response positions by normalized full-vocabulary
student entropy,
\[
s^{\mathrm{Entropy}}=\widetilde H.
\]
The selector score is computed entirely from the student.

\paragraph{TIP.}
TIP uses the top-$K$ reverse-KL proxy for $D_{\mathrm{KL}}(P_S\|P_T)$
provided by OPD, computed as the negative sum of its vocabulary-wise rewards.
Full-vocabulary student entropy is upper-clipped at the batch 98th percentile
and min--max normalized to $\widehat H$; the proxy is separately min--max
normalized to $\widehat d$. TIP ranks positions using the Soft-OR score
\[
s^{\mathrm{TIP}}
=
\widehat H+\widehat d-\widehat H\widehat d.
\]

\paragraph{TA-OPD.}
Here $D$ is the forward teacher-to-student KL approximation over the union of
student and teacher top-$K$ supports. Each distribution is renormalized on the
union, treating missing entries as zero, then floored at $10^{-12}$ without
a second renormalization.
The reported TA-OPD baseline corresponds to \texttt{dlearn\_high} in the
implementation and uses
\[
s^{\mathrm{TA}}
=
\widetilde D\,\widetilde C_{\mathrm{mass}}.
\]
The compatibility signal is intended to measure teacher probability mass on
the student top-$K$ support. The reported configuration returns the teacher's
own top-$K$ distribution, so the implemented proxy sums teacher
probability mass over the student-teacher top-$K$ intersection.

\subsection{Counterfactual and Control Selectors}

\paragraph{CF-OPD.}
CF-OPD holds the sampled response fixed and measures the teacher distribution
shift between the original and counterfactual prompts,
\[
s_{i,t}^{\mathrm{CF}}
=\operatorname{JSD}_{K}
\!\left(P_{i,t}^{o,T},P_{i,t}^{c,T}\right).
\]
Here, $P_{i,t}^{o,T}$ and $P_{i,t}^{c,T}$ denote the fixed teacher
distributions obtained by rescoring the sampled response under the original
and counterfactual prompts.
It uses the same top-$K$ union and residual-probability bucket as CROP, with
$K=16$ in our experiments. CF-OPD consumes the counterfactual prompt and uses
a distribution-level JSD score based on the contrast between the original and
counterfactual prompts.

\paragraph{PC-OPD.}
PC-OPD is the meaning-preserving paraphrase ablation of CROP. It uses
the same top-$K$
JSD-with-residual approximation between the original and paraphrased teacher
distributions and assigns
\[
s_{i,t}^{\mathrm{PC}}
=
-\operatorname{JSD}_{K}
\!\left(P_{i,t}^{o,T},P_{i,t}^{p,T}\right).
\]
Here, $P_{i,t}^{p,T}$ denotes the fixed teacher distribution obtained by
rescoring under the meaning-preserving paraphrase.
Because all selectors retain high-ranking scores, the negative sign favors
positions that remain stable under the meaning-preserving paraphrase.

\paragraph{CROP-student.}
\label{app:crop-student}
CROP-student retains the student-side scoring rule used in the corresponding
ablation. For the fixed on-policy response
$y_i\sim\pi_{\bar\theta}(\cdot\mid x_i)$, it evaluates the same response token
and prefix under the three matched prompts using the frozen rollout-time
student:
\[
P_{i,t}^{o,S}=\pi_{\bar\theta}(\cdot\mid x_i,y_{i,<t}),\quad
P_{i,t}^{p,S}=\pi_{\bar\theta}(\cdot\mid x_i^{\mathrm{para}},y_{i,<t}),\quad
P_{i,t}^{c,S}=\pi_{\bar\theta}(\cdot\mid x_i^{\mathrm{cf}},y_{i,<t}).
\]
Each distribution is represented by its top-$K$ vocabulary entries, with
$K=16$ in the reported experiment. CROP-student uses the same union support,
residual-probability bucket, and Jensen--Shannon divergence from
Equations~\ref{eq:crop-residual-support}--\ref{eq:crop-jsd}, and assigns
\[
s_{i,t}^{\mathrm{CROP\text{-}student}}
=\operatorname{JSD}_{K}(P_{i,t}^{o,S},P_{i,t}^{c,S})
-\operatorname{JSD}_{K}(P_{i,t}^{o,S},P_{i,t}^{p,S}).
\]
The score is converted into a binary mask using the shared per-response token budget
in Equation~\ref{eq:crop-response-budget}, and selected tokens use the unchanged
sampled-token OPD loss in Equation~\ref{eq:crop-loss}. Thus, CROP-student and
CROP share the rollout, score form, budget rule, and OPD target. CROP-student
sources the matched selector distributions from the student, and CROP sources
them from the teacher. In CROP-student, the student computes the selector score
and the teacher supplies the OPD target.

\paragraph{Random.}
Random assigns independent pseudorandom scores to valid response positions
and applies the shared per-response token budget. The seed is the configured base
seed plus the global training step, making each step reproducible while
changing the mask across steps.

\paragraph{CROP-Bottom.}
CROP-Bottom reverses the teacher-scored CROP ranking and retains the
lowest-scoring positions within each response under the same token budget. It
is a matched control that measures the effect of ranking direction under a
fixed per-response token budget.

\begin{table}[t]
\centering
\caption{Comparison of teacher versus student rescoring across nominal per-response token
budgets in Setting A (Avg., \%). CROP uses matched teacher rescoring, whereas
CROP-student applies the same score form and budget rule with student
rescoring. $\Delta$ denotes CROP minus CROP-student.}
\label{tab:teacher-student-budget}
\small
\setlength{\tabcolsep}{4.2pt}
\begin{adjustbox}{max width=\columnwidth}
\begin{tabular}{lcccc}
\toprule
Method & 5\% & 10\% & 15\% & 20\% \\
\midrule
CROP-student & 26.08 & 25.92 & 25.22 & 26.37 \\
CROP & \textbf{26.47} & \textbf{26.11} & \textbf{26.13} & \textbf{26.98} \\
$\Delta$ & $+0.39$ & $+0.19$ & $+0.91$ & $+0.61$ \\
\bottomrule
\end{tabular}
\end{adjustbox}
\end{table}

Across the four evaluated budgets, CROP consistently outperforms
CROP-student, with gains ranging from 0.19 to 0.91 points. This extends the
20\% scorer ablation and shows that the advantage of teacher rescoring holds
throughout the tested 5--20\% budget range in Setting A.

\subsection{Selected-token Category Analysis}
\label{app:selected-token-categories}

We analyze the token positions selected by Entropy, TIP, TA-OPD, and CROP on
the same Setting A responses under the matched 20\% per-response budget. We
decode the exported byte-level BPE representation with the Qwen3-1.7B
tokenizer before classification, preserving spaces, line breaks, full-width
symbols, and other surface forms used by the categorization rules.

Each selected token is assigned to exactly one of five categories using the
fixed precedence order \emph{task-specific} $\rightarrow$
\emph{formatting/punctuation} $\rightarrow$ \emph{boilerplate reasoning}
$\rightarrow$ \emph{function words} $\rightarrow$ \emph{content words}.
Task-specific tokens include numbers, mathematical operators, single-letter
variables, Greek letters, and mathematical concepts such as
\texttt{sum}, \texttt{equation}, and \texttt{probability}.
Formatting/punctuation includes punctuation marks, brackets, line breaks,
special tokens, and \LaTeX{} delimiters. Boilerplate reasoning includes generic
reasoning templates such as \texttt{let}, \texttt{therefore}, \texttt{thus},
\texttt{we have}, \texttt{answer}, \texttt{first}, \texttt{next}, and
\texttt{finally}; multi-token templates are matched on the decoded local token
sequence, and their constituent tokens receive the same category. Function
words include articles, prepositions, conjunctions, pronouns, and auxiliary
verbs. All remaining ordinary lexical tokens are assigned to content words.
The precedence rule resolves overlaps and ensures that every selected token is
counted once.

\subsection{Entropy-Augmented CROP}
\label{app:crop-ent}

CROP-ent augments the teacher-sourced CROP relevance score with normalized
student entropy. The entropy component is computed from the student top-$16$
rollout distribution, while the relevance component uses the same top-$K$
matched teacher rescoring as CROP. Applying the shared batch normalization in
Equation~\ref{eq:selector-batch-normalization} gives
$\widetilde H^K_{i,t}=\operatorname{Norm}_{\mathcal B}(H(P_{i,t}^{o,S})/\log K)$ and
$\widetilde s_{i,t}^{\mathrm{CROP}}=
\operatorname{Norm}_{\mathcal B}(s_{i,t}^{\mathrm{CROP}})$. CROP-ent combines
the two normalized signals with a Soft-OR score:
\begin{equation}
s_{i,t}^{\mathrm{CROP\text{-}ent}}
=\widetilde H^K_{i,t}+\widetilde s_{i,t}^{\mathrm{CROP}}
-\widetilde H^K_{i,t}\widetilde s_{i,t}^{\mathrm{CROP}}.
\label{eq:crop-ent-score}
\end{equation}
Equation~\ref{eq:crop-ent-score} is used for ranking before the shared
per-response token budget and minimum-one rule are applied. Selection remains
independent within each response.

\begin{table}[t]
\centering
\caption{CROP and CROP-ent in Setting A across nominal retention ratios
(Avg., \%). CROP uses the teacher-sourced relevance score, and CROP-ent adds
normalized student entropy to that score. Bold marks the higher value at each
ratio.}
\label{tab:crop-ent-budget}
\small
\setlength{\tabcolsep}{4.2pt}
\begin{adjustbox}{max width=\columnwidth}
\begin{tabular}{lccccc}
\toprule
Method & 10\% & 20\% & 30\% & 40\% & 50\% \\
\midrule
CROP & \textbf{26.11} & \textbf{26.98} & \textbf{25.79} & 24.64 & \textbf{24.98} \\
CROP-ent & 23.75 & 24.04 & 24.91 & \textbf{25.00} & 24.31 \\
\bottomrule
\end{tabular}
\end{adjustbox}
\end{table}

Table~\ref{tab:crop-ent-budget} isolates the interaction between task
relevance and entropy. At 10\% and 20\%, CROP exceeds CROP-ent by 2.37 and
2.94 points, respectively. Task relevance therefore has the stronger effect
when the available supervision is sparse. The gap narrows to 0.89 points at
30\% and reverses at 40\%, where CROP-ent leads by 0.36 points. At 50\%, CROP
again leads by 0.67 points. Entropy therefore becomes more competitive as the
budget grows and is more effective at 40\%, but this benefit does not persist
at the largest tested budget. The results support a robust role for task
relevance at low budgets and a localized high-budget benefit from entropy.

\subsection{Implementation Summary}

All methods operate on the same on-policy student responses. Pure OPD retains
the standard dense objective, IW-OPD continuously weights its token advantages,
and Dense CREDIT replaces the teacher correction with a contrastive signal at
every valid position. The hard selectors preserve the standard sampled-token
OPD signal and construct a per-response mask from a method-specific ranking
score. Their selector-side scoring procedures are method-specific. All methods
share the student rollout procedure.

Table~\ref{tab:selector-implementation} summarizes the score source,
selector-side inputs or rescoring, ranking score, and top-$K$ configuration of
each method.

\begin{table*}[t]
\centering
\caption{Implementation-level comparison of the OPD methods. Dense methods use
all valid response positions; hard selectors use the shared per-response token
budget in Equation~\ref{eq:crop-response-budget}.}
\label{tab:selector-implementation}
\small
\setlength{\tabcolsep}{4pt}
\begin{adjustbox}{max width=\textwidth}
\begin{tabular}{lllll}
\toprule
Method & Mode & Signal source & Rule & $K$ \\
\midrule
Base
& untrained
& --
& student initialization
& -- \\

Pure OPD
& dense
& student + teacher
& standard OPD on all valid positions
& -- \\

IW-OPD
& dense
& sampled student--teacher gap
& continuous weight $w_{i,t}$, $w_{\max}=1.5$
& -- \\

Dense CREDIT
& dense
& teacher + one negative rescore
& contrastive teacher correction, $\lambda=0.1$
& 16 \\

Entropy
& hard selector
& student
& full-vocabulary entropy
& -- \\

TIP
& hard selector
& student + teacher
& $\widehat H+\widehat d-\widehat H\widehat d$
& 16 (KL proxy) \\

TA-OPD
& hard selector
& student + teacher
& $\widetilde D\,\widetilde C_{\mathrm{mass}}$
& 16 \\

CF-OPD
& hard selector
& teacher
& counterfactual JSD
& 16 \\

PC-OPD
& hard selector
& teacher
& negative paraphrase JSD
& 16 \\

Random
& hard selector
& seeded random scores
& random ranking within response
& -- \\

CROP-Bottom
& hard selector
& teacher
& lowest CROP scores
& 16 \\

CROP-student
& hard selector
& student
& counterfactual JSD minus paraphrase JSD
& 16 \\

CROP-ent
& hard selector
& student + teacher
& Soft-OR of normalized student entropy and CROP
& 16 \\

CROP
& hard selector
& teacher
& counterfactual JSD minus paraphrase JSD
& 16 \\
\bottomrule
\end{tabular}
\end{adjustbox}
\end{table*}

\subsection{Training Hyperparameters and Hardware}
\label{app:training-hyperparameters}

Each complete training batch contains 144 prompts with four student responses
per prompt. Training runs for 115 steps with a fixed data order, using maximum
prompt and response lengths of 1,024 and 4,096 tokens. The objective is
\texttt{token\_reward\_direct}; all runs use $\lambda_{\mathrm{OPD}}=1.0$ and
PPO clipping radii $(\epsilon,\epsilon_{\mathrm{high}})=(0.2,0.28)$. The
optimizer is AdamW with learning rate $10^{-6}$, betas $(0.9,0.999)$, weight
decay 0.01, a constant schedule from the first step, gradient clipping at 1.0, and
BF16 precision. Losses are aggregated with token mean.
All experiments were run on a single machine equipped with eight NVIDIA
GeForce RTX 5090 GPUs.

Entropy and TIP use full-vocabulary student entropy; TIP's OPD disagreement
proxy and TA-OPD use top-$16$ quantities. CF-OPD, PC-OPD, CROP-Bottom,
and CROP use matched top-$K$ teacher rescoring; only CROP-student uses matched
top-$K$ student rescoring among these ablations. CROP-ent combines top-$16$
student rollout entropy with the top-$K$ CROP relevance signal from matched
teacher rescoring. Apart from TIP's clipped-entropy/min--max normalization,
batch-normalized selector metrics use the 5--95\% quantile transform in
Equation~\ref{eq:selector-batch-normalization}.

All hard selectors independently apply the same rule for computing the
per-response token budget and the same minimum-one safeguard. Consequently,
the realized retention ratio may lie
slightly below or above the nominal retention ratio $\rho$.

Both settings use the same scientific training configuration. Differences in
GPU placement, offload, dynamic token limits, and memory utilization serve as
engineering accommodations for model size. The data, objective, batch
semantics, optimizer, horizon, and training steps remain identical.

\subsection{Budgeted CREDIT Control}
\label{app:budgeted-credit}

To test whether applying a sparse token budget to CREDIT closes the gap to
CROP, we additionally evaluate Budgeted CREDIT in Setting A. This variant
ranks valid response positions by the CREDIT contrastive score and retains the
top 20\% independently within each response. It uses one negative prompt,
$\lambda=1.0$, and negative-selection seed 42. Because the Dense CREDIT
baseline uses $\lambda=0.1$, this comparison evaluates the resulting budgeted
variant and does not isolate the effect of sparsity alone.

\begin{table}[htbp]
\centering
\caption{Budgeted CREDIT control in Setting A (\%). Benchmark entries report
Mean/Pass with eight samples for AIME24, AIME25, and AMC23 and four samples for
the 492-problem MATH-500 clean subset. Dense CREDIT and CROP are repeated from
Table~\ref{tab:main_results}. Budgeted CREDIT uses evaluation seeds 1000--1007
for AIME24/AIME25/AMC23 and 5000--5003 for MATH-500. Bold marks the best Avg.}
\label{tab:budgeted-credit}
\small
\begin{adjustbox}{max width=\linewidth}
\begin{tabular}{llccccc}
\toprule
Method & Budget & AIME24 & AIME25 & AMC23 & MATH-500 & Avg. \\
\midrule
Dense CREDIT & 100\% & 4.17 / 20.00 & 4.17 / 16.67 & 34.06 / 65.00 & 58.69 / 75.20 & 25.27 \\
Budgeted CREDIT & 20\% & 4.17 / 16.67 & 5.42 / 20.00 & 32.19 / 62.50 & 59.45 / 79.07 & 25.31 \\
CROP & 20\% & 8.33 / 23.33 & 3.75 / 10.00 & 35.00 / 67.50 & 60.82 / 78.86 & \textbf{26.98} \\
\bottomrule
\end{tabular}
\end{adjustbox}
\end{table}

Budgeted CREDIT reaches 25.31 Avg., effectively matching Dense CREDIT at
25.27 but remaining 1.67 points below CROP. Thus, the observed Budgeted CREDIT
variant does not close the aggregate gap to CROP under a matched 20\% nominal
token budget.

\section{Evaluation Details}
\label{app:evaluation-protocol}

Table~\ref{tab:evaluation-settings} summarizes the decoding configurations for
the four-benchmark evaluation. All runs use the same rule-based mathematical
answer grader and require the final answer in
\texttt{\textbackslash boxed\{\}} form.
Each trained checkpoint is evaluated with multiple seeds for sampled
decoding. The reported Mean@8/Pass@8 and Mean@4/Pass@4 aggregate these
evaluations of a fixed checkpoint; they do not estimate variability across
independent training runs.

\paragraph{Software environment.}
The evaluation environment uses Python~3.12.13, PyTorch~2.8.0,
Transformers~4.57.1, and vLLM~0.11.0. The answer grader uses
SymPy~1.14.0 and pylatexenc~2.11 for symbolic checking and LaTeX parsing.

\begin{table}[htbp]
\centering
\caption{Evaluation decoding configurations. A/A/A denotes AIME24, AIME25,
and AMC23.}
\label{tab:evaluation-settings}
\small
\begin{adjustbox}{max width=\linewidth}
\begin{tabular}{llccccc}
\toprule
Setting & Benchmarks & Thinking & Samples & Temp. & Top-$p$ & Max tokens \\
\midrule
A & A/A/A & yes & 8 & 0.7 & 0.95 & 31,744 \\
A & MATH-500 & yes & 4 & 0.7 & 0.95 & 31,744 \\
B & A/A/A & no & 8 & 0.7 & 0.95 & 31,744 \\
B & MATH-500 & no & 4 & 0.7 & 0.95 & 31,744 \\
\bottomrule
\end{tabular}
\end{adjustbox}
\end{table}

\subsection{Generation Diagnostics for the Main Results}

Table~\ref{tab:main-generation-diagnostics} reports available generation
statistics for the comparison checkpoints associated with
Table~\ref{tab:main_results}.
These diagnostics help identify format and length effects, but are not used to
attribute the performance differences to a particular mechanism.

\begin{table*}[t]
\centering
\caption{Available generation diagnostics for the comparison runs in
Table~\ref{tab:main_results}. Train Trunc. is the fraction of responses
reaching the maximum training response length at the final step. Avg. Len. and
Format Fail. are measured on the reported evaluation generations; Format Fail.
denotes outputs without a parsable \texttt{\textbackslash boxed\{\}} answer. The
final MATH-500 column is the fraction of generations terminated by the
evaluation length limit. All MATH-500 evaluation diagnostics use the same
492-problem clean subset as Table~\ref{tab:main_results}.}
\label{tab:main-generation-diagnostics}
\scriptsize
\setlength{\tabcolsep}{2.8pt}
\begin{adjustbox}{max width=\textwidth}
\begin{tabular}{
llcr
rrrrrrrrr
}
\toprule
Setting & Method & Budget & Train Trunc.
& \multicolumn{2}{c}{AIME24}
& \multicolumn{2}{c}{AIME25}
& \multicolumn{2}{c}{AMC23}
& \multicolumn{3}{c}{MATH-500} \\
\cmidrule(lr){5-6}\cmidrule(lr){7-8}\cmidrule(lr){9-10}\cmidrule(lr){11-13}
& & & & Avg. Len. & Format Fail. & Avg. Len. & Format Fail.
& Avg. Len. & Format Fail. & Avg. Len. & Format Fail. & Trunc. \\
\midrule
\multirow{8}{*}{\makecell[l]{Setting A\\4B $\rightarrow$ 1.7B}}
& Base & -- & --
& 5,415.05 & 18.75\% & 5,642.47 & 19.58\% & 2,939.12 & 10.94\%
& 1,885.93 & 9.96\% & 4.32\% \\
& Pure OPD & 100\% & 4.69\%
& 8,833.12 & 30.00\% & 4,464.05 & 14.58\% & 3,559.97 & 11.88\%
& 1,900.13 & 7.52\% & 4.32\% \\
& IW-OPD & 100\% & 5.03\%
& 7,346.83 & 23.33\% & 7,144.06 & 23.33\% & 3,396.06 & 10.63\%
& 2,027.21 & 7.83\% & 4.83\% \\
& Dense CREDIT & 100\% & 4.51\%
& 7,380.64 & 22.08\% & 5,160.22 & 16.67\% & 2,804.63 & 8.13\%
& 1,941.79 & 7.57\% & 4.52\% \\
\cmidrule(lr){2-13}
& Entropy & 20\% & 5.21\%
& 6,365.79 & 20.83\% & 7,182.20 & 25.42\% & 3,458.26 & 10.63\%
& 2,120.27 & 7.88\% & 5.03\% \\
& TIP & 20\% & 6.25\%
& 6,580.79 & 20.42\% & 5,755.18 & 20.83\% & 3,785.12 & 12.50\%
& 2,011.58 & 7.98\% & 4.73\% \\
& TA-OPD & 20\% & 7.29\%
& 8,232.55 & 26.25\% & 4,872.35 & 15.42\% & 3,274.50 & 9.38\%
& 2,119.89 & 7.72\% & 5.03\% \\
\cmidrule(lr){2-13}
& CROP & 20\% & 6.94\%
& 6,860.83 & 20.00\% & 4,816.31 & 17.08\% & 4,031.19 & 11.88\%
& 2,296.52 & 8.18\% & 5.44\% \\
\midrule
\multirow{8}{*}{\makecell[l]{Setting B\\8B $\rightarrow$ 4B}}
& Base & -- & --
& 4,569.88 & 4.17\% & 3,230.47 & 0.42\% & 1,899.63 & 0.63\%
& 930.26 & 0.00\% & 0.00\% \\
& Pure OPD & 100\% & 1.03\%
& 5,720.37 & 4.17\% & 4,435.48 & 1.67\% & 2,024.70 & 0.00\%
& 1,093.17 & 0.20\% & 0.36\% \\
& IW-OPD & 100\% & 1.22\%
& 6,134.13 & 6.25\% & 4,896.40 & 4.17\% & 1,840.56 & 0.00\%
& 1,097.35 & 0.36\% & 0.46\% \\
& Dense CREDIT & 100\% & 1.22\%
& 5,485.40 & 3.33\% & 4,391.42 & 3.33\% & 2,089.55 & 0.00\%
& 1,130.18 & 0.41\% & 0.61\% \\
\cmidrule(lr){2-13}
& Entropy & 20\% & 1.56\%
& 6,503.11 & 8.75\% & 4,674.75 & 3.33\% & 1,831.06 & 0.00\%
& 1,076.62 & 0.41\% & 0.46\% \\
& TIP & 20\% & 1.22\%
& 5,976.08 & 7.08\% & 4,485.05 & 2.92\% & 1,894.51 & 0.31\%
& 1,033.03 & 0.10\% & 0.20\% \\
& TA-OPD & 20\% & 1.39\%
& 5,875.83 & 5.42\% & 4,684.37 & 3.33\% & 1,976.66 & 0.31\%
& 1,180.81 & 0.46\% & 0.71\% \\
\cmidrule(lr){2-13}
& CROP & 20\% & 1.39\%
& 5,898.52 & 7.50\% & 4,899.90 & 4.58\% & 2,128.29 & 0.00\%
& 1,123.57 & 0.25\% & 0.46\% \\
\bottomrule
\end{tabular}
\end{adjustbox}
\end{table*}

\subsection{Training--Evaluation Contamination Audit}
\label{app:contamination-audit}

\paragraph{Scope and protocol.}
We audit overlap between the task-specific distillation data and all four
mathematical evaluations. The audit covers all 16,594 validated training
instances and all three prompt fields---the original $x$, paraphrase
$x^{\mathrm{para}}$, and counterfactual $x^{\mathrm{cf}}$---for a total of
49,782 prompt texts. Exact matching canonicalizes Unicode, whitespace, and
presentational \LaTeX{} while retaining numbers and mathematical operators.
We then retrieve token-trigram near-duplicate candidates and manually verify
whether each candidate expresses the same mathematical problem or differs only
in a material condition. This manual step prevents shared problem templates
from being counted as leakage.

\paragraph{Overlap findings.}
Table~\ref{tab:contamination-summary} summarizes the audit, and
Table~\ref{tab:contamination-ids} lists the verified MATH-500 pairs. We find no
confirmed exact or near-duplicate overlap with AIME24, AIME25, or AMC23. Seven
unique MATH-500 problems exactly match original training prompts; these
correspond to eight training records because one problem occurs twice. No exact
match occurs only through a generated paraphrase or counterfactual field. We
additionally find one condition-level near duplicate: MATH-500 contains
$z^4+z^2+1=0$, whereas the corresponding training prompt contains
$z^4-z^2+1=0$. We conservatively exclude it because the two problems differ by
only one mathematical sign.

\begin{table}[t]
\centering
\caption{Training--evaluation overlap after manual verification. The clean
subset excludes the union of exact and condition-level near duplicates.}
\label{tab:contamination-summary}
\small
\begin{tabular}{lrrrr}
\toprule
Benchmark & Size & Exact & Near duplicate & Clean size \\
\midrule
AIME24 & 30 & 0 & 0 & 30 \\
AIME25 & 30 & 0 & 0 & 30 \\
AMC23 & 40 & 0 & 0 & 40 \\
MATH-500 & 500 & 7 & 1 & 492 \\
\bottomrule
\end{tabular}
\end{table}

\begin{table*}[t]
\centering
\caption{Verified MATH-500 overlaps. The nested-radical problem appears in two
training records, so seven exact benchmark overlaps correspond to eight
training records. The last row is conservatively treated as a near duplicate.}
\label{tab:contamination-ids}
\scriptsize
\setlength{\tabcolsep}{4pt}
\begin{tabular}{@{}>{\raggedright\arraybackslash}p{0.30\textwidth}>{\raggedright\arraybackslash}p{0.27\textwidth}>{\raggedright\arraybackslash}p{0.35\textwidth}@{}}
\toprule
MATH-500 ID & Training instance ID & Problem identifier \\
\midrule
\path{test/intermediate_algebra/1849.json} & \texttt{dapo\_math\_011108} & $\log(kx)=2\log(x+2)$ \\
\path{test/intermediate_algebra/582.json} & \texttt{dapo\_math\_004487} & roots of $x^{10}+(13x-1)^{10}=0$ \\
\path{test/intermediate_algebra/991.json} & \texttt{dapo\_math\_007920} & $3m+4n=100$; minimize $|m-n|$ \\
\path{test/algebra/1282.json} & \texttt{dapo\_math\_013144}, \texttt{dapo\_math\_016696} & integer values of $\sqrt{120-\sqrt{x}}$ \\
\path{test/intermediate_algebra/1411.json} & \texttt{dapo\_math\_006264} & positive-integer roots of a cubic polynomial \\
\path{test/intermediate_algebra/90.json} & \texttt{dapo\_math\_004530} & fourth-order alternating recurrence \\
\path{test/intermediate_algebra/2146.json} & \texttt{dapo\_math\_009074} & bounded quadratic polynomial $P(x)$ \\
\midrule
\path{test/precalculus/285.json} & \texttt{dapo\_math\_003831} & $z^4+z^2+1$ versus $z^4-z^2+1$ (near) \\
\bottomrule
\end{tabular}
\end{table*}

\paragraph{Decontaminated evaluation.}
We form \textsc{MATH-500-Clean} by removing the eight affected evaluation
problems. The MATH-500 results reported in this paper are recomputed on this
492-problem subset from the retained per-example outputs; no new generations
or selective filtering by model outcome are performed. The reported aggregate
scores likewise use \textsc{MATH-500-Clean}, and CROP retains its aggregate
lead after decontamination. This audit addresses contamination introduced by
the task-specific distillation set; it cannot establish which public benchmark
material may have appeared during base-model pretraining.

\section{Qualitative Analysis of the CROP Signal}
\label{app:crop-qualitative-analysis}

\begin{figure*}[!t]
\centering
\includegraphics[width=0.96\textwidth]{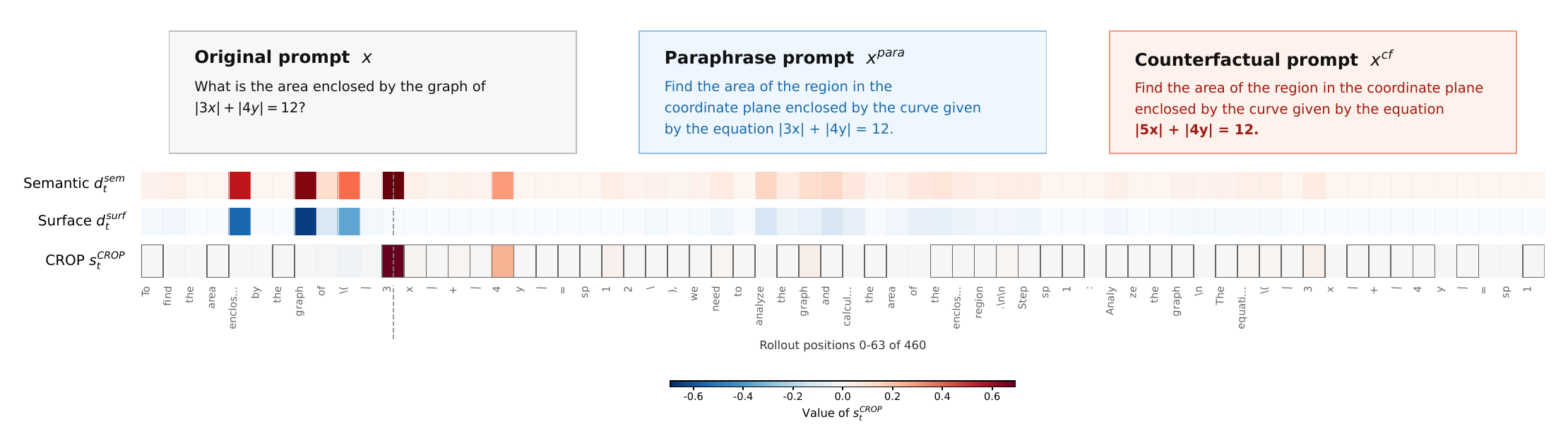}
\par\vspace{3pt}
\includegraphics[width=0.96\textwidth]{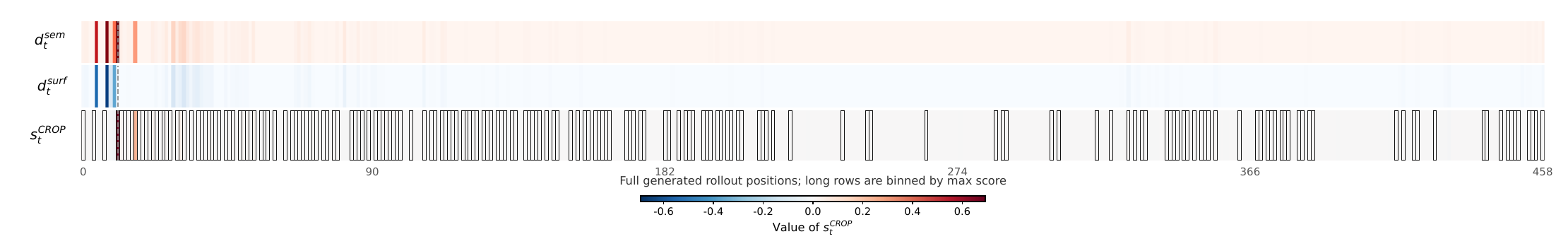}
\caption{CROP diagnostic for Case A, where the counterfactual changes the
coefficient of $|x|$ from $3$ to $5$. Top: the local 64-position peak window.
Bottom: the complete rollout overview.}
\label{fig:crop-case-a}
\end{figure*}

\begin{figure*}[!t]
\centering
\includegraphics[width=0.96\textwidth]{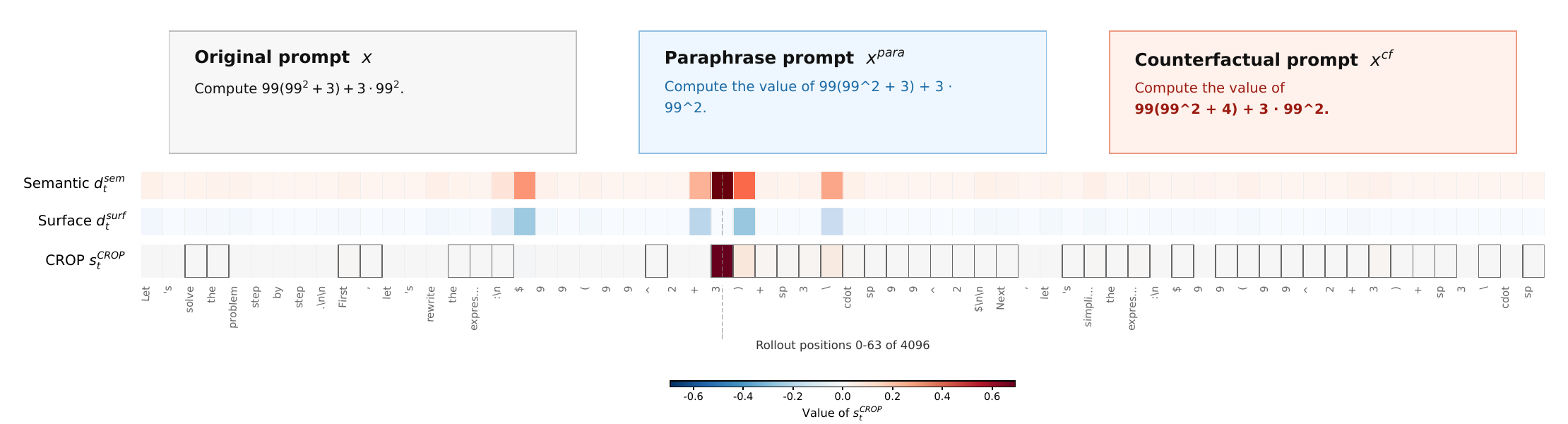}
\par\vspace{3pt}
\includegraphics[width=0.96\textwidth]{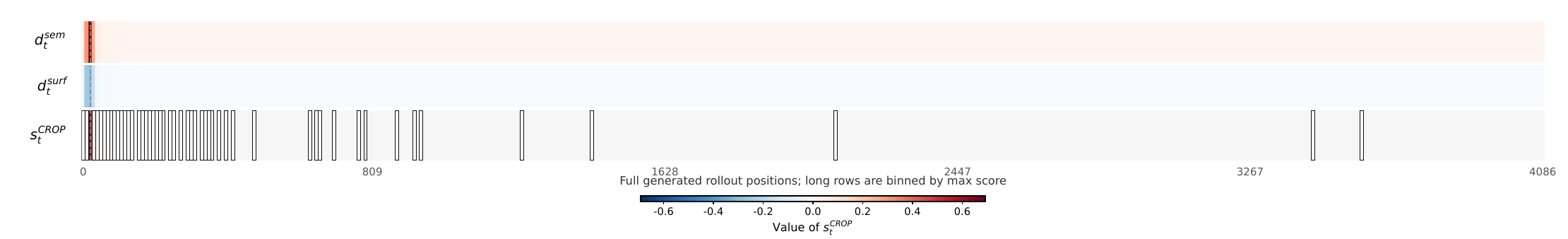}
\caption{CROP diagnostic for Case B, where the counterfactual changes the
explicit additive constant in the $99$-power expression from $3$ to $4$.
Top: the local 64-position peak window. Bottom: the complete rollout overview.}
\label{fig:crop-case-b}
\end{figure*}

\begin{figure*}[!t]
\centering
\includegraphics[width=0.96\textwidth]{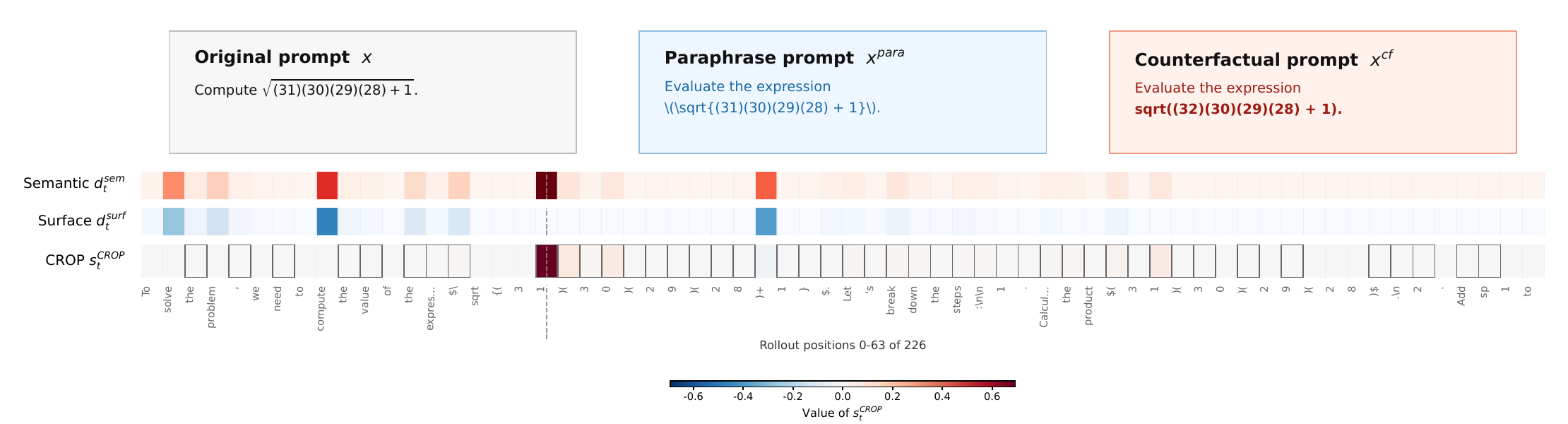}
\par\vspace{3pt}
\includegraphics[width=0.96\textwidth]{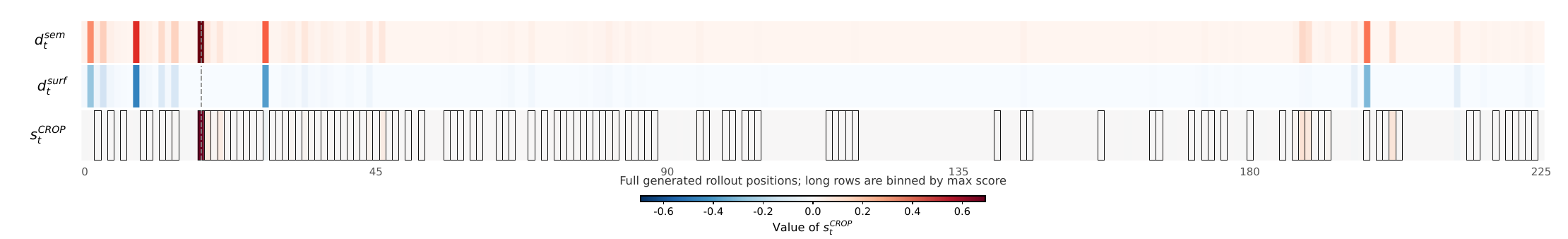}
\caption{CROP diagnostic for Case C, where the counterfactual changes $31$ to
$32$ in the radical expression. Top: the local 64-position peak window.
Bottom: the complete rollout overview.}
\label{fig:crop-case-c}
\end{figure*}

\begin{figure*}[!t]
\centering
\includegraphics[width=0.96\textwidth]{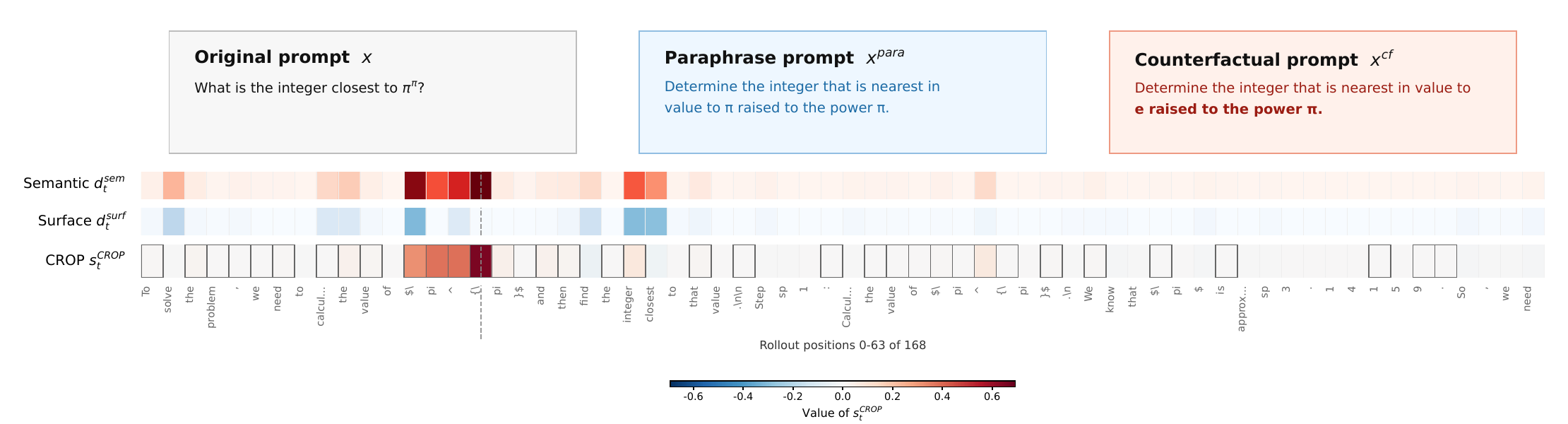}
\par\vspace{3pt}
\includegraphics[width=0.96\textwidth]{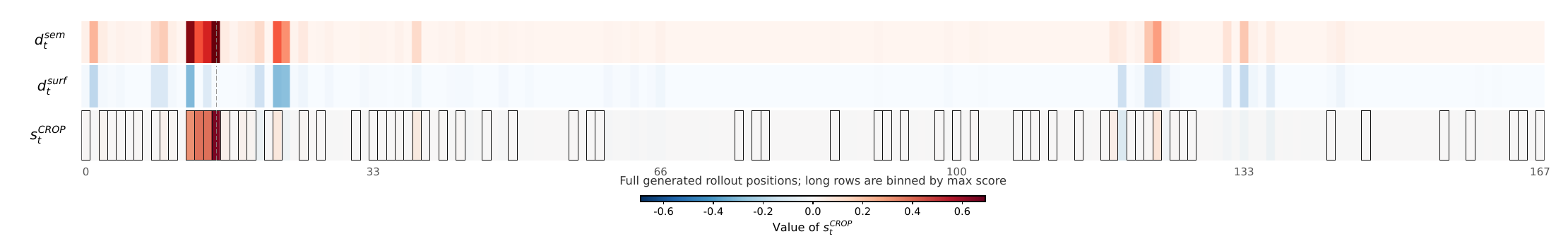}
\caption{CROP diagnostic for Case D, where the counterfactual replaces the
base $\pi$ by $e$. Top: the local 64-position peak window. Bottom: the full
rollout overview.}
\label{fig:crop-case-d}
\end{figure*}

Figures~\ref{fig:crop-case-a}--\ref{fig:crop-case-d} show four exact
token-level diagnostics. For each example, the upper panel expands a
64-position window containing the largest CROP score, while the lower panel
places the same signal in the complete fixed rollout. The heatmaps separate
semantic sensitivity, surface sensitivity, and their CROP margin; outlined
positions are retained by the CROP mask. These examples illustrate how a
localized counterfactual response can coexist with selections elsewhere in a
long rollout. They are qualitative mechanism checks, not an estimate of
corpus-wide localization accuracy.

\FloatBarrier

\section{Reproducibility Details}
\label{app:reproducibility-details}

All methods use the same 16,594-record data snapshot. After prompt-length filtering, 16,588 records remain, of which 16,560 are used to form 115 complete batches of 144 prompts; the remaining 28 records are omitted to keep all training batches complete. Every prompt produces four responses. Each method rolls out
from the original prompt; selectors read paraphrase or counterfactual metadata
only when their definitions require it. Main selective comparisons use a 20\%
per-response token budget unless another budget is explicitly shown.

We do not estimate variability across independent training seeds.
Nevertheless, CROP outperforms the strongest matched-budget non-CROP hard
selector at all four budgets in Setting A and leads the strongest non-CROP
baseline in both teacher--student settings at the main 20\% budget,
suggesting that its advantage is not confined to a single experimental
configuration.

\section{Limitations and Future Work}
\label{app:limitations}

Our experiments focus on two Qwen teacher--student configurations and
mathematical reasoning tasks. Evaluating CROP across additional model
families and domains is an important direction for future work.

CROP relies on automatically constructed paraphrase--counterfactual triplets
and their validation quality. Our validation pipeline explicitly checks
semantic preservation, single-condition intervention, determinacy, and
answer-type consistency.
Future work can further study alternative intervention-generation strategies,
multiple valid interventions per prompt, and extensions to coding and
open-domain reasoning.

\end{document}